\documentclass[sigconf]{acmart}
\AtBeginDocument{%
  }

\usepackage{enumitem}
\usepackage{booktabs}
\usepackage{arydshln}
\usepackage{bbm}
\usepackage{varwidth}
\usepackage{wrapfig}
\usepackage{placeins}
\setcopyright{acmlicensed}
\copyrightyear{2018}
\acmYear{2018}
\acmDOI{XXXXXXX.XXXXXXX}
\acmConference[Conference acronym 'XX]{Make sure to enter the correct
  conference title from your rights confirmation email}{June 03--05,
  2018}{Woodstock, NY}
\acmISBN{978-1-4503-XXXX-X/2018/06}

\begin{document}

\title[DELUGE: Towards Continental-Scale Daily Pluvial Flood Damage Prediction]{DELUGE: Towards Continental-Scale Daily Pluvial Flood Damage Prediction via Interpretable Conditioning on Foundation Model Embeddings}

\author{Yuya Kawakami}
\affiliation{%
  \institution{University of California, Davis}
  \city{Davis}
  \state{California}
  \country{USA}
}

\author{Daniel Cayan}
\affiliation{%
  \institution{Scripps Institution of Oceanography\\University of California, San Diego}
  \city{La Jolla}
  \state{California}
  \country{USA}
}

\author{Dongyu Liu}
\affiliation{%
  \institution{University of California, Davis}
  \city{Davis}
  \state{California}
  \country{USA}
}

\author{Kwan-Liu Ma}
\affiliation{%
  \institution{University of California, Davis}
  \city{Davis}
  \state{California}
  \country{USA}
}

\author{Tom Corringham}
\affiliation{%
  \institution{Scripps Institution of Oceanography\\University of California, San Diego}
  \city{La Jolla}
  \state{California}
  \country{USA}
}


\renewcommand{\shortauthors}{Kawakami et al.}
\newcommand{\fix}[1]{%
    \colorbox{red!20}{\textbf{Fix:} #1}%
}
\newcommand{\colorchip}[1]{\textcolor[HTML]{#1}{\rule[-0ex]{1.5ex}{1.5ex}}}

\newcommand{\shortnote}[1]{%
    \colorbox{blue!20}{\begin{varwidth}{\dimexpr\linewidth-2\fboxsep\relax}\textbf{Note:} #1\end{varwidth}}%
}

\newcommand{\note}[1]{%
    \fcolorbox{blue!50}{blue!20}{\parbox{\dimexpr\columnwidth-2\fboxsep-2\fboxrule}{\textbf{Note:} #1}}%
}
\newcommand{\tsim}[1]{${\sim}}

\setlength{\intextsep}{2pt}
\setlength{\textfloatsep}{2pt}
\setlength{\abovecaptionskip}{2pt}
\setlength{\belowcaptionskip}{2pt}


\begin{abstract}

Pluvial (rainfall-driven) flooding accounts for 45\% of National Flood Insurance Program (NFIP) claims in the United States and is harder to predict than its riverine and coastal counterparts, with existing approaches limited to coarse resolution, regional domains, or computationally intensive process-based models unsuitable for daily continental-scale use.
%
% We present \textbf{DELUGE}, a multimodal deep learning framework for daily pluvial flood damage prediction across the Conterminous United States (CONUS) at ${\sim}1$\,km resolution, trained on spatially and temporally corrected NFIP claims (2017--2022) and structured around the hazard, exposure, and vulnerability components of disaster risk.
We present \textbf{DELUGE}, a multimodal deep learning framework for daily pluvial flood damage prediction at ${\sim}1$\,km resolution and national scale, trained on spatially and temporally corrected NFIP claims (2017--2022) and structured around the hazard, exposure, and vulnerability components of disaster risk.
Rather than blanket coverage of the Conterminous United States (CONUS), we model the top 100 highest-claim 75\,km cells, distributed nationwide and accounting for \tsim$81\%$ of total pluvial flood claims.
Our architectural novelty is a pair of parametric modules in the hydrometeorology branch, a \emph{Value Modulator} and a \emph{Temporal Modulator}, conditioned on terrain descriptors and AlphaEarth foundation-model embeddings, that expose directly inspectable hydrological response parameters and provide architecture-level \emph{interpretability-by-design}.
Under a spatial block holdout, DELUGE outperforms tuned Random Forest, XGBoost, and LightGBM baselines by 9\% to 30\% on a dollar-weighted area under the precision-recall curve (PR-AUC), a metric that emphasizes the rare, high-cost claims of greatest operational interest.
Beyond DELUGE, we argue this interpretable conditioning scheme is a transferable pattern for integrating foundation-model embeddings into other geospatial prediction tasks.
%
% To our knowledge, this is the first high-resolution, continental-scale system for daily pluvial flood damage prediction in the United States, enabling nationwide analyses relevant to risk management, insurance, and large-scale flood impact research.
% %
% Beyond DELUGE itself, the modulator design offers a concrete pattern for incorporating Earth-observation foundation-model embeddings into downstream prediction with interpretability in mind.
\end{abstract}

%-----




\received{20 February 2007}
\received[revised]{12 March 2009}
\received[accepted]{5 June 2009}

\maketitle
% \vspace{-11pt}

\section{Introduction}
Flooding is among the costliest natural hazards in the conterminous United States (CONUS)~\cite{fema_cost_of_flooding}, with damage dynamics that vary sharply by topography, drainage infrastructure, and climate regime~\cite{brunner_spatial_2020}.
Among flood types, \emph{pluvial} flooding (inundation driven by intense rainfall exceeding local drainage capacity) is a particularly difficult prediction target~\cite{rosenzweig_pluvial_2018,nelson-mercer_pluvial_2025}.
%
% Events are highly localized in space and time, develop on hourly timescales, and account for roughly $45\%$ of National Flood Insurance Program (NFIP) claims and approximately \$56~billion of insured damage since 1980~\cite{fema_nfip_claims_v2}.
%
Anticipating pluvial damage at a resolution operationally meaningful to insurers, emergency managers, and municipal planners is both an open scientific problem and is  of immediate societal value.

Despite this urgency, pluvial flood damage prediction lags behind its riverine and coastal counterparts~\cite{rosenzweig_pluvial_2018}.
Riverine flooding benefits from a dense network of stream gauges that provide a relatively clean, continuous training signal~\cite{boutayeb_when_2025,nearing_global_2024}, and coastal flooding is similarly well-tracked by tide and surge gauges~\cite{bates_combined_2021,yang_predicting_2022}.
Pluvial flooding, on the other hand, has a noisy observational backbone, and its damage signal lives in scattered insurance claims~\cite{fema_nfip_claims_v2} and citizen reports~\cite{puttinaovarat_flood_2020,mayo_groundsource_2026}, each carrying nontrivial spatial and temporal uncertainty.
CONUS-scale pluvial products do exist, but they occupy a different operating regime than ours.
FEMA flood maps~\cite{fema_flood_zone_2021} are static delineations of long-term flood hazard zones rather than dynamic estimates of damage, while CONUS-scale hydrodynamic models~\cite{wing_30_2024,bates_combined_2021} simulate physical inundation at high fidelity but are computationally heavy and produced for fixed design scenarios rather than on a daily basis.
Neither delivers a \emph{lightweight, daily, ${\sim}1$\,km prediction of insured pluvial damage}, the regime we target.

We close this gap with \textbf{DELUGE} (\textit{\textbf{D}}aily \textit{\textbf{E}}stimation of p\textit{\textbf{LU}}vial flood dama\textit{\textbf{GE}}), to our knowledge the first end-to-end learned system to predict daily, ${\sim}1$\,km insured pluvial flood damage across the highest-claim regions of CONUS through interpretable conditioning on Earth-observation foundation-model embeddings.
DELUGE is a multimodal CNN whose local inductive bias matches the locally driven nature of pluvial dynamics, in contrast to the catchment-scale connectivity behind GNN-based riverine approaches~\cite{sarkar_hydrogat_2025,kazadi_floodgnn-gru_2024}.
We train DELUGE on NFIP claims as a damage proxy, after applying temporal and spatial uncertainty corrections that refine each claim's timing and footprint.
Under a spatial block holdout protocol that partitions 75\,km grid cells into a train/test split, DELUGE achieves a PR-AUC of \tsim$0.24$, well above the \tsim$0.0025$ no-skill baseline implied by the task's \tsim$0.25\%$ positive rate, and outperforms tuned Random Forest, XGBoost, and LightGBM baselines commonly used in claims-based flood-damage prediction, with the largest gains on high-cost claims.
% (standard in the pluvial flood literature) by 9\% to 26\% on PR-AUC and 11\% to 35\% on a dollar-weighted variant.

Beyond its predictive performance, DELUGE's architectural novelty lies in two parametric modules within the hydrometeorology branch.
The \emph{Value Modulator} learns a per-sample monotone warp of each hydrometeorological input, and the \emph{Temporal Modulator} learns a per-sample temporal-response kernel that causally convolves the hydrometeorological time series.
Both modules are conditioned on local place characteristics, encoded through terrain descriptors and Google AlphaEarth foundation-model embeddings~\cite{brown_alphaearth_2025}, which supply a dense, learned representation of the built and natural environment that varies continuously across CONUS.
% and captures local context that hand-crafted terrain descriptors alone miss.

Rather than concatenating these embeddings into a black-box encoder, we route them through the physics-shaped modulators, whose parameters carry direct hydrological meaning such as response peak lag in hours and decay timescale.
This makes the modulators a principled and interpretable way to consume an otherwise opaque foundation embedding, yielding \emph{interpretability-by-design}~\cite{rudin_stop_2019} where ConvLSTM or Transformer-style encoders would require post-hoc attribution approximations like SHAP~\cite{lundberg_unified_2017}.
Thus our interpretability is at the architecture level, exposing the per-location hydrological response DELUGE has learned, rather than a per-prediction explanation.
Our main contributions are:
\begin{enumerate}[nosep, leftmargin=14pt]
    \item \textbf{DELUGE}, the first end-to-end learned system for daily pluvial flood damage prediction at ${\sim}1$\,km across the highest-claim regions of CONUS, outperforming tree-based baselines standard in the pluvial flood prediction literature, with the largest gains on high-cost claims.
    \item Two uncertainty-correction procedures for NFIP claims, a precipitation guided correction of claim dates and a spatial refinement that intersects redacted geographies to localize damage to higher resolution.
    \item The \emph{Value Modulator} and \emph{Temporal Modulator}, terrain- and AlphaEarth-conditioned parametric modules whose learned per-patch parameters are themselves hydrologically meaningful, yielding \emph{interpretability-by-design} of Geospatial Foundation Model integration.
    \item Inspection of the learned modulator parameters via geography-blind clustering, showing they recover hydrologically coherent regimes consistent with physical expectation.
    % \item Empirical verification of the modulators' physical fidelity via geography-blind clustering that recovers physically grounded explanation of identified regimes.
\end{enumerate}
\vspace{-5pt}

% DELUGE shows that with appropriate parametric inductive biases, sparse and noisy flood damage labels are enough to recover physically meaningful structure, supporting future data-driven approaches to continental-scale flood impact modeling.

\section{Related Works}
\paragraph{Flood Hazard Modeling and Process-Based Approaches}
Flood risk mapping has traditionally relied on process-based hydrological and hydraulic models that simulate fluid physics over high-resolution topography~\cite{kumar_comprehensive_2023,bates_simple_2000}.
These models underpin FEMA flood maps and CONUS-scale ambient-flood products~\cite{fema_flood_zone_2021,bates_combined_2021,woznicki_development_2019,wing_30_2024}, providing 30\,m-to-street-level inundation footprints valuable for insurance and long-term planning, often at decadal scales.
For damage estimation specifically, the insurance industry has developed catastrophe (CAT) models that integrate hazard, exposure, and vulnerability for detailed risk assessment~\cite{zanardo_introduction_2022}.
However, neither of these approaches is suited to daily, dynamic damage prediction. 
Ambient-flood products characterize long-run risk rather than the daily prediction, and CAT models are computationally expensive to run at scale~\cite{bates_flood_2022,kazadi_floodgnn-gru_2024}.

\paragraph{Machine Learning for Flood Prediction}
Machine learning approaches address the operational gap left by process-based models by learning directly from large archives of hydrological, topographical, and socioeconomic data~\cite{oddo_deep_2024,collins_predicting_2022,alipour_leveraging_2020,liao_fast_2023}.
Tabular tree-based methods (Random Forest, XGBoost, LightGBM) are the dominant choice in the flood damage prediction literature~\cite{yang_predicting_2022,collins_predicting_2022,garcia_reconstructing_2025,alipour_leveraging_2020}, typically operating on aggregated features at the county or grid-cell level.
A parallel line of work applies sequence and graph models (LSTM, GNN) to riverine and flash-flood prediction~\cite{nearing_global_2024,tran_ai_2025,kazadi_floodgnn-gru_2024,sarkar_hydrogat_2025,taghizadeh_interpretable_2024,moishin_designing_2021}, achieving strong results, but the targets are typically river gauge measurements or hazard indicators rather than insured damage directly.
A further body of remote-sensing work maps flood extent from satellite imagery~\cite{rambour_sen12-flood_2020,Bonafilia_2020_CVPR_Workshops,puttinaovarat_flood_2020}, but likewise recovers inundation footprints rather than damage.
Among damage-targeted ML works, most are constrained to regional domains (a single county, or city) where local data permit detailed modeling~\cite{liu_flooddamagecast_2024,zhang_high_2023,shu_assessing_2026}, leaving the CONUS-wide daily pluvial damage problem largely unaddressed.

\paragraph{Pluvial Flood Damage Prediction from Claims}
Pluvial flooding is a particularly difficult prediction target~\cite{rosenzweig_pluvial_2018,nelson-mercer_pluvial_2025}, due to the highly localized natures of impacts and lack the dense observational infrastructure of stream and tide gauges that supports riverine and coastal flood predictions.
NFIP claims are the most comprehensive source of insured pluvial damage in the United States~\cite{fema_nfip_claims_v2} and have become the de facto damage signal for data-driven CONUS-scale work~\cite{yang_predicting_2022,garcia_reconstructing_2025}.
However, claims data are known to carry nontrivial spatial and temporal uncertainty.
Data redaction to protect personally identifiable information limits localization to Census Tract or ZIP polygons, reported dates frequently misalign with precipitation events due to reporting lags and mislabeling~\cite{wing_new_2020,shin_systematic_2022,petkov_learning_2025},
and uninsured losses are systematically absent (\tsim$2/3$ of total losses~\cite{amornsiripanitch2024flood}).
Past work has largely treated these uncertainties as fixed properties of the dataset. 
Here, we instead build an explicit temporal and spatial uncertainty correction pipeline (\S\ref{ssec:temporal_correction},~\S\ref{ssec:spatial_correction}) to refine each claim before it enters the model, which we view as a prerequisite for using NFIP claims as a daily training signal.

\paragraph{GeoAI Foundation Models and Interpretability}
Earth-observation foundation models are increasingly becoming backbones of geospatial analysis tasks~\cite{agarwal2024general,szwarcman2024geospatial,zhu_foundations_2026}.
Efforts like Google AlphaEarth~\cite{brown_alphaearth_2025} provide dense, high-resolution learned embeddings that implicitly encode the built and natural environment and spatial context.
A growing body of work also probes \emph{what} these embeddings represent about physical space~\cite{benavides2026earth,rahman_physically_2026,bell_earth_2026}.
However, less attention has been paid to \emph{how} downstream models should consume them.
The default pattern, which is to concatenate embeddings into a black-box encoder, makes it difficult to verify whether the geospatial prior is being used in a physically meaningful way.
This raises the same interpretability concerns that motivate broader calls for explainable GeoAI~\cite{hsu_explainable_2023,roussel2025introducing,suri_trusting_nodate,zhu_foundations_2026} and interpretable flood models~\cite{taghizadeh_interpretable_2024}, and that Rudin~\cite{rudin_stop_2019} has addressed by advocating \emph{inherently} interpretable architectures over post-hoc attribution.
DELUGE's interpretable conditioning modules are designed to fill exactly this gap.
Rather than treating foundation-model embeddings as opaque feature vectors, we route them through parametric, physics-shaped modules whose learned parameters are themselves hydrologically meaningful and inspectable.

\vspace{-0.09in}
\section{Data}
\subsection{Prediction Target: FIMA NFIP Claims}
Our prediction target is derived from the OpenFEMA FIMA NFIP Redacted Claims Dataset v2~\cite{fema_nfip_claims_v2}, which records every NFIP insurance claim filed under the National Flood Insurance Program, including the reported \texttt{dateOfLoss}, redacted location (Census Tract, ZIP code, and nearest $0.1^{\circ}\times 0.1^{\circ}$ grid cell), \texttt{causeOfDamage} category, and paid amounts for building and contents coverage.
As we target pluvial claims, we filter the dataset to include only those claims defined as \texttt{causeOfDamage} = \texttt{Accumulation of rainfall or snowmelt}, which serve as the supervisory signal for our prediction task.
NFIP data, while the most comprehensive source of insured flood damage in CONUS, is known to have uncertainties and errors in both space and time~\cite{wing_new_2020,petkov_learning_2025,shin_systematic_2022}.
To improve learning dynamics and model validation, we apply two correction methods, temporal and spatial, to mitigate these uncertainties.

% \subsubsection{Temporal and Spatial Uncertainty Correction}
% %
% Such data noise pose challenges in effective learning. 
% %
% To improve learning dynamics and model validation, we target two sources of uncertainty in the NFIP claims data: temporal and spatial.
% %
\subsubsection{Temporal Uncertainty Correction}\label{ssec:temporal_correction}
The \texttt{dateOfLoss} field in the NFIP claims data is known to be noisy where reported dates of loss often fail to align with precipitation events in the historical precipitation records.
These discrepancies may arise from reporting lags, human error during claim filing, and misclassification of \texttt{causeOfDamage}, where fluvial or coastal damages are mislabeled as pluvial~\cite{shin_systematic_2022}.
To mitigate this noise, we apply the following correction to the \texttt{dateOfLoss} field $d$:
\begin{enumerate}[nosep, align=left, leftmargin=*, left=1pt]
    \item Compute the daily maximum hourly accumulated precipitation from NOAA AORC~\cite{fall_analysis_2023} for day $d$ at the location of the claim.
    \item If the daily maximum hourly accumulated precipitation on day $d$ exceeds 10\,mm, accept \texttt{dateOfLoss} as reported.
    \item Otherwise, search the window $d' \in \{d-3, \ldots, d+3\}$ and select the day with the largest daily maximum hourly precipitation. If this maximum exceeds 10\,mm, update \texttt{dateOfLoss} to $d'$.
    \item If no day's daily maximum hourly precipitation in the window exceeds 10\,mm, discard the claim as erroneous.
\end{enumerate}

\noindent
We adopt 10\,mm as a \textit{conservative} threshold for maximum hourly accumulated precipitation, sitting between two reference points. 
National Weather Service Flash Flood Guidance (FFG) thresholds are most predictive for most areas when hourly precipitation exceeds \tsim$25.4$\,mm~\cite{james_precipitation_2024}, while the American Meteorological Society's Glossary of Meteorology defines \textit{heavy} precipitation as 7.62\,mm in one hour~\cite{ams_glossary_rain}.
%
% Table~\ref{tab:date_shifts} summarizes the resulting shift distribution. 
With this processing, $81.1\%$ of claims are accepted as reported, $12.9\%$ are shifted within $\pm 3$ days, and $6.0\%$ are discarded.
We place the full table of date shifts in the Appendix.

\subsubsection{Spatial Uncertainty Correction}\label{ssec:spatial_correction}
The address-level claims data is not publicly available and the redacted claims data is only identifiable by its Census Tract, Zip Code and the nearest $0.1^{\circ}\times 0.1^{\circ}$ degree cell.
%
% This spatial uncertainty in the exact location is problematic for learning.
To mitigate this spatial uncertainty of the location of the claim, for every claim, we retrieve the intersection of the reported Census Tract, Zip Code, and the $0.1^{\circ}\times 0.1^{\circ}$ cell centered on the reported coordinate, yielding a polygon for each claim.
After this spatial uncertainty correction, our median polygon area for each claim is $\sim1.7$km$^2$ with a mean area of $\sim7.7$km$^2$ due to a significant right tail of large rural polygons.

\subsection{Study Region and Period}\label{ssec:study_region}
\begin{figure}[t]
    \centering
    \includegraphics[width=\linewidth]{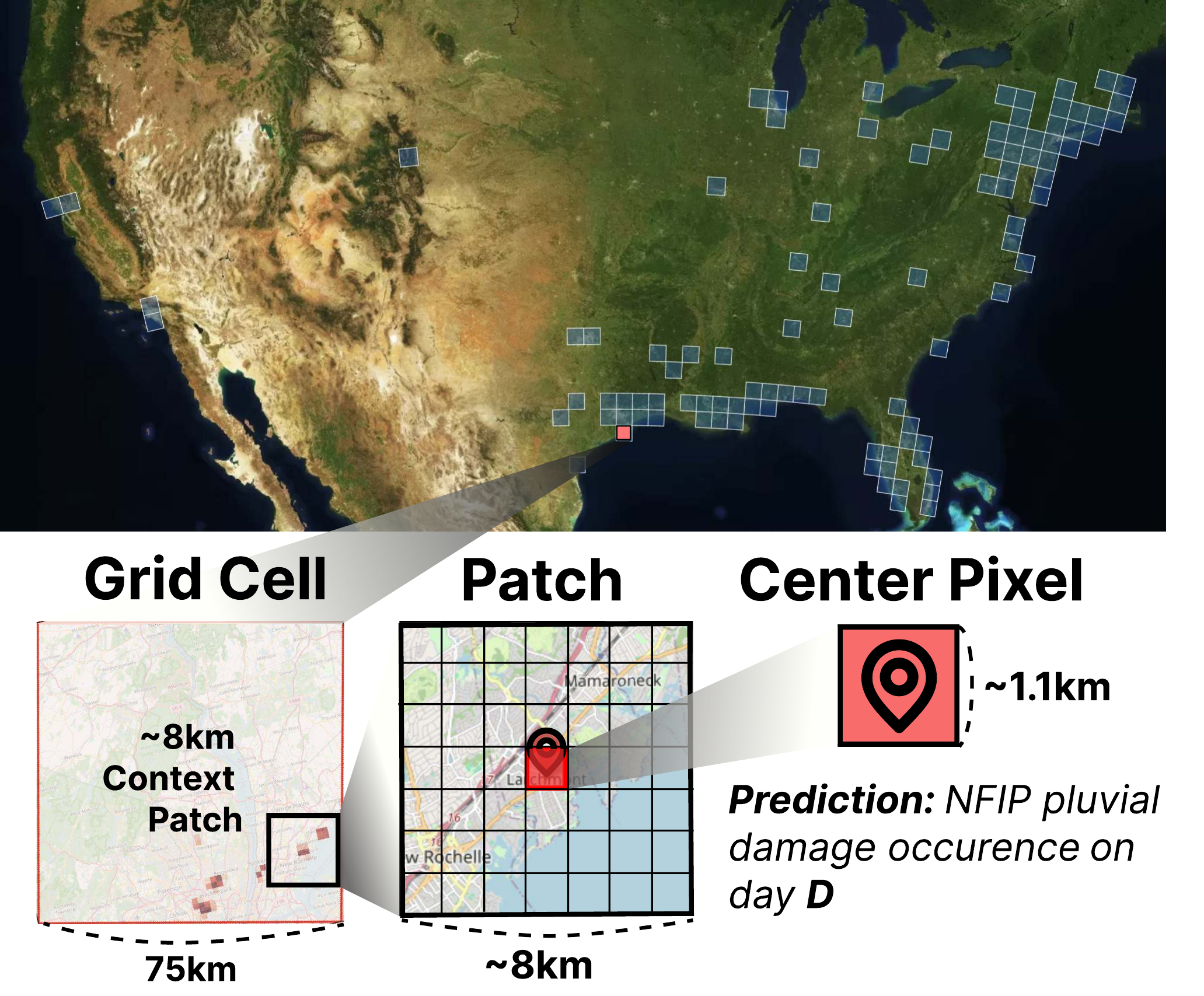}
    \caption{Our study region in blue. We select the Top 100 75$\times$75\,km \emph{grid cells} for NFIP pluvial flood damage claims, accounting for $\sim$81\% of claims from 2017--2022.  After rasterizing NFIP claims (see \S\ref{ssec:temporal_correction},\S\ref{ssec:spatial_correction}), training samples take the form of (\emph{pixel}, day) tuples, and DELUGE takes input a $\sim$8x8km \emph{patch} centered around the \emph{pixel} as spatial context.}
    \label{fig:study_region}
\end{figure}

We segment CONUS into 75$\times$75\,km grid cells and select the top 100 cells by pluvial NFIP damage claim count, yielding a nationally distributed set of the most flood-affected regions rather than blanket coverage of CONUS.
These cells span CONUS, while concentrating in the Northeast and Gulf regions of the United States (Fig.~\ref{fig:study_region}), and together account for $\sim$81\% of all pluvial NFIP claims from 2017--2022.
Extending to the top 200 cells would raise claim coverage only modestly to $\sim$91\% while roughly doubling the dataset, so we adopt the top 100 as a balance between capturing the large majority of pluvial claims and keeping storage and compute tractable.
Within each 75$\times$75\,km grid cell, we further discretize space into a 64$\times$64 raster of \textit{pixels} ($\sim$1.17\,km on a side), which constitute the unit of spatial prediction throughout the remainder of the paper.
We restrict our study period to 2017--2022, the longest contiguous interval over which all data sources have full-year coverage.
%
% This window is bounded below by the AlphaEarth Foundation embedding record, which is available annually from 2017, and above by the NWM Retrospective v3.0 dataset, which ends in January 2023.

\subsection{Dataset Construction}\label{ssec:dataset_construction}
The unit of binary prediction is a (pixel, day) tuple over our study region and period, i.e., was there a pluvial NFIP damage claim reported at pixel $p$ on day $d$?
First to collect our positive class, we rasterize each claim's spatial-uncertainty polygon (\S\ref{ssec:spatial_correction}) onto the 64$\times$64 grid, labeling every intersecting pixel as positive.
To prevent claims with large polygons from dominating the loss, we associate each claim with a confidence score: $c_{claim} = \min(1,\, A_{\text{pix}}/A_{\text{claim}})$ with $A_{\text{pix}} = 1.17^{2}$\,km$^{2}$.
To account for pixels with many intersecting claims, we calculate and define our confidence at each pixel $c_{pixel}$ as the complement of the product of each claim's confidence score: $c_{pixel} = 1 - \prod_{i\in\text{claim}}(1 - c_i)$.
This operation captures the intuition that multiple claims intersecting a pixel should increase the confidence of a claim for that pixel.
% while also preventing pixels with many intersecting claims from dominating the loss by effectively
% scaling each claim's total contribution toward that of a single-pixel-localized claim.

For the negative class, exhaustively enumerating every pixel-day in our study period would yield a label distribution dominated by trivially negative dry days, further exacerbating class imbalance.
We therefore construct the dataset along three data types drawn from two contrastive axes: positives, \textit{spatial} negatives (same-day, different-pixel), and \textit{temporal} negatives (different-day, same-grid).
% \begin{itemize}[noitemsep, topsep=0pt, left=0pt, align=left]    
%
    \textbf{Positives}: \emph{Pixel-days containing at least one NFIP pluvial claim after the temporal and spatial uncertainty corrections described above, along with a confidence score $c$.}
    \textbf{Spatial Negatives}: \emph{Pixel-days without a claim on grid-days that contain at least one claim somewhere within the same 75$\times$75\,km grid cell.}
    \textbf{Temporal Negatives}: \emph{Pixel-days on grid-days where the spatial mean of the 1-hour maximum accumulated precipitation from NOAA AORC exceeds 10\,mm and no NFIP claim was filed anywhere in the grid cell.} 
We reuse the 10\,mm threshold from our temporal uncertainty correction (\S\ref{ssec:temporal_correction}) so that meaningful precipitation events are defined consistently across the pipeline.
% \end{itemize}
Despite this undersampling of temporal negatives, the dataset exhibits \emph{significant class imbalance} with a positive rate of \tsim$0.25\%$ across the study region and period.

% \section{Methods}
% \input{tex/methods.tex}
\section{DELUGE Architecture}
\begin{figure*}
    \centering
    \includegraphics[width=\textwidth]{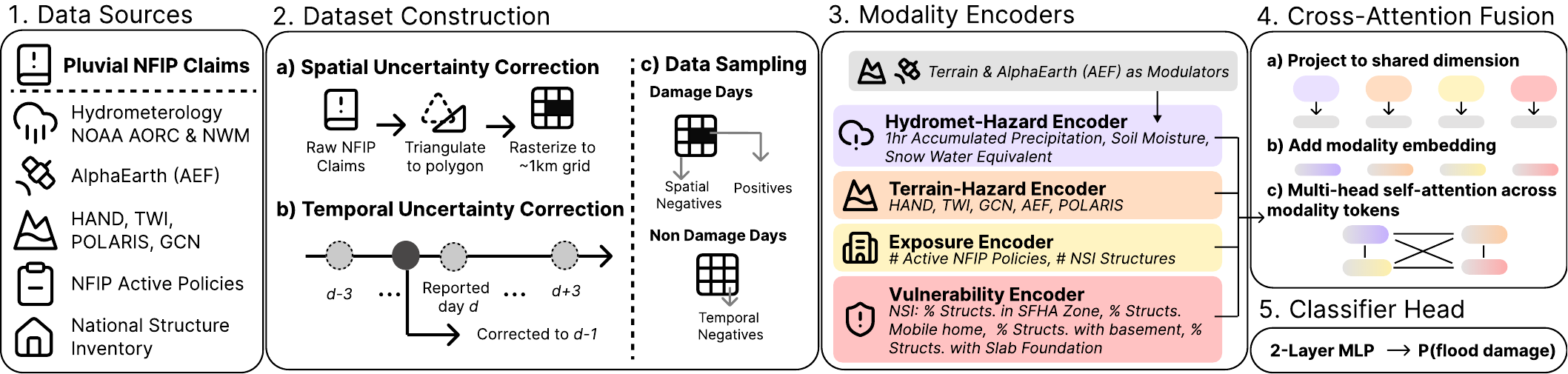}
    \caption{\textbf{DELUGE architecture.} DELUGE is a multimodal CNN-based model that integrates various data modalities, including meteorological, terrain-hazard, vulnerability, and exposure features, to predict pluvial flood damage at a daily and $\sim$1km resolution across the highest-claim regions of CONUS. The model consists of separate CNN branches for each modality, which are then fused together to produce the final damage prediction. The hydrometeorology branch is described in
 detail in Fig.~\ref{fig:modulators}.}
    \vspace{-0.15in}
    \label{fig:deluge_architecture}
\end{figure*}
Pluvial flooding is often a localized event with dynamics that depend strongly on local conditions such as topography, land use, and drainage infrastructure.
Although pluvial events, especially street-level events, may occur at sub-kilometer scales~\cite{lowe_urban_2020} beyond the resolution of our modeling, we provide the model with a larger \tsim$8\times8$\,km spatial context centered on each sample (a $7\times7$ pixel patch) for two reasons.

First, prior work has identified the spatial scale at which pluvial flooding develops~\cite{cristiano_spatial_2017,weiler_pluvial_2025}.
The effective extent varies across locations, but our \tsim$8\times8$\,km context covers the scale of pluvial flooding identified in the literature.
Second, pluvial flood events are often driven by convective storms~\cite{li_conterminous_2022} that deposit large amounts of precipitation in a localized area.
These extreme precipitation events are less reliably captured in the NOAA AORC archive~\cite{fall_analysis_2023}, and we expect their reported locations to carry variable spatial noise even with the latest observational and radar-based products.
A larger spatial context lets the model associate observed damage with the precipitation pattern in the surrounding area, which is more reliably recorded than the storm's exact center.
% with the spatial scale of pluvial flooding and associated hydrological process identified in the literature \cite{cristiano_spatial_2017}.
%
These considerations motivate DELUGE, the multimodal CNN-based model described in the following sections.
\vspace{-5pt}
  \subsection{Overview}
\label{sec:arch-overview}

We predict per-pixel binary pluvial flood insurance claims occurrence at $\sim$1.17~km resolution across the highest-claim regions of CONUS (\S\ref{ssec:study_region}) by fusing four modality branches (hydrometeorology, terrain, exposure, and vulnerability) into a per-pixel probability (Figure~\ref{fig:deluge_architecture}).
With respect to the standard triad in disaster risk modeling, the hydrometeorology and terrain branches capture the hazard, while the exposure and vulnerability branches capture exposure and vulnerability respectively.
Our target is the center pixel of a $7 \times 7$ patch (\tsim$8\times8$km) and all branches operate on this patch.
% Under the severe class imbalance of NFIP claims ($\sim$0.25\% positive, $\sim$1{,}800 events in $\sim$720K samples), we partition training, validation, and test cells via AEF-informed spatial clustering (Section~\ref{sec:arch-training}) to prevent geographic leakage.
\subsubsection{Supporting interpretability}
Predictive accuracy alone is insufficient for models intended to inform real-world flood-risk decisions.
Insurers and emergency managers must be able to inspect and justify a model's reasoning before acting on its outputs, especially under the high financial stakes of flood damage.
Prior work in deep-learning based flood and hydrological modeling has emphasized this same need for interpretability~\cite{taghizadeh_interpretable_2024, li_interpretable_2024, lee_improving_2024}.
We target this need at the architecture level, designing modulators (\S\ref{sec:arch-warp} and \ref{sec:arch-kernel}) whose hydrological response parameters are directly inspectable, a distinct goal from per-prediction interpretability.

\begin{figure*}[ht]
    \centering
    \includegraphics[width=\textwidth]{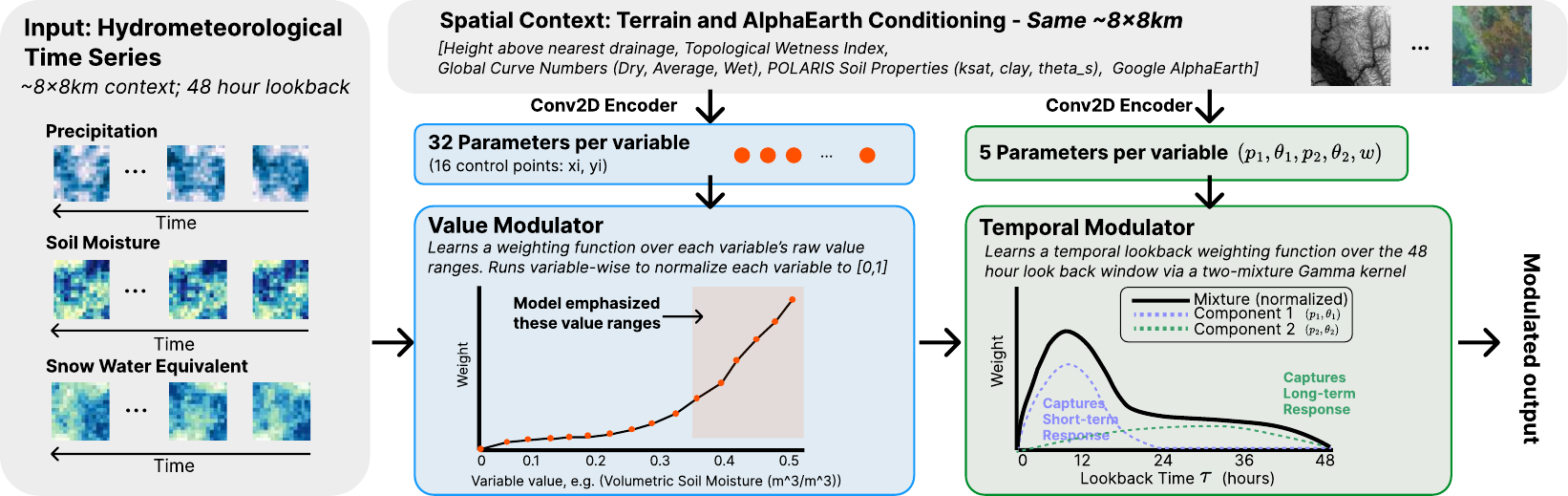}
    \caption{The two terrain- and AEF-conditioned modulators acting on the hydrometeorology branch. The \textbf{Value Modulator} (\S\ref{sec:arch-warp}) applies a per-channel, per-patch piecewise-linear warp $\phi_{v,i}:[0,1]\!\to\![0,1]$ to each min-max normalized input, reshaping the response curve along the value axis. The \textbf{Temporal Modulator} (\S\ref{sec:arch-kernel}) then convolves the warped sequence with a per-channel gamma-kernel mixture $k_v(\tau)$,
    integrating the $T$-hour lookback along the time axis. Each modulator has its own conditioning network (ConvBlock + MLP) over the stacked terrain and AlphaEarth Foundation features, making the learned warp shapes and kernel peaks/decay timescales directly interpretable per patch.}\label{fig:modulators}
    \vspace{-0.15in}
\end{figure*}

We expose this interpretability via the hydrometeorology branch, the most important modality for prediction, where we develop two terrain and foundation model-conditioned modules that modulate the hydrometeorology signal along its two axes.
% \begin{itemize}[noitemsep, topsep=0pt, left=0pt, align=left]    
    % \item \emph{Value modulator} (Section~\ref{sec:arch-warp}), which learns a per-patch, per-channel monotonic warp of the hydrometeorology value range (value axis).
    % \item \emph{Temporal modulator} (Section~\ref{sec:arch-kernel}), which learns a per-patch mixed gamma-kernel hydrological response over the hydrometeorology lookback window (time axis); and
% \end{itemize}
First, the \emph{Value modulator} (Section~\ref{sec:arch-warp})  learns a per-patch, per-channel monotonic warp of the hydrometeorology value range (value axis) and then the \emph{Temporal modulator} (Section~\ref{sec:arch-kernel}), which learns a per-patch mixed gamma-kernel hydrological response over the hydrometeorology lookback window (time axis).
Both modules are terrain and foundation model-conditioned, so their parameters vary spatially as explicit model outputs (Fig.~\ref{fig:modulators}).
Conditioned on the local place characteristics, they answer two questions: which range of values in the hydrometeorology signal is most relevant to flood damage (\emph{Value Modulator}), and what temporal pattern of the hydrometeorology signal is most relevant to flood damage (\emph{Temporal Modulator}).

A similar module to our \emph{Value Modulator} was offered in \cite{taghizadeh_interpretable_2024} via Kolmogorov-Arnold Networks (KAN), but here we allow the model to learn location-specific warps that are conditioned on each local terrain and foundation model features.
This is a deliberate \emph{interpretability-by-design} choice rather than the more common path of pairing a flexible black-box encoder (ConvLSTM, Transformer) with post-hoc attribution (SHAP, integrated gradients, attention visualization)~\cite{rudin_stop_2019}, allowing us to inspect the model's reasoning directly from hydrologically meaningful, spatially explicit modulator parameters instead of generic feature importance scores.

\vspace{-7pt}
\subsection{Data Inputs}\label{ssec:arch-inputs}
% Each sample is a $7 \times 7$ pixel patch centered on the prediction pixel, plus higher-resolution modality features. Climate is sampled hourly over a $T = 73$~hour lookback ending at the prediction date $D$; terrain is static; vulnerability and AEF are annual (Table~\ref{tab:inputs}).

Predicting flood damages requires accounting for the three components of risk, hazard, exposure, and vulnerability, and we gather a comprehensive set of inputs to capture each.
\textbf{Hydrometeorological hazard} is drawn from hourly NOAA AORC accumulated precipitation~\cite{fall_analysis_2023} and 3-hourly NOAA National Water Model Retrospective volumetric soil moisture and snow water equivalent~\cite{cosgrove_noaas_2024}.
\textbf{Terrain hazard} combines the Topographic Wetness Index~\cite{hoylman_30m_2021}, Height Above Nearest Drainage~\cite{aws_copernicus_dem}, Global Curve Number under three antecedent runoff conditions~\cite{jaafar_gcn250_2019}, NLCD fractional impervious surface~\cite{usgs_annual_nlcd_2024}, and POLARIS soil properties (saturated hydraulic conductivity, clay percentage, and saturated water content)~\cite{chaney_polaris_2019}.
\textbf{Exposure and vulnerability} come from FIMA NFIP active policy counts and coverage~\cite{fema_nfip_policies_v2} and per-structure attributes from the USACE National Structure Inventory~\cite{usace_nsi_2022} (structure count, foundation types, mobile-home share, Special Flood Hazard Area [SFHA] membership, first-floor elevation statistics).
Finally, to complement these and to condition our modulators, we include 64-dimensional \textbf{Google AlphaEarth Foundation (AEF)} embeddings at 10\,m resolution~\cite{brown_alphaearth_2025}, which implicitly encode land cover and the built environment.
Per-source descriptions, exact variables used, and feature normalization details are deferred to the Appendix.

For each sample, a pixel-day tuple, we extract data for our modalities from the $7 \times 7$ pixel patch centered on the prediction pixel, bilinearly interpolating to match our $\sim$1.17~km resolution regular grid.
For inputs that are natively offered at a higher resolution, all sources except for the hydrometeorology and NFIP active policies, we extract the data at $56\times 56$ resolution.
For our hydrometeorology inputs, we extract a $T=73$-hour input window spanning Day~$D-2$ through Day~$D$ (inclusive).
The temporal modulator (\S\ref{sec:arch-kernel}) convolves this window with an $L=48$-hour kernel, and the temporal reduction selects a peak-response hour within Day~$D$.
The two are sized so that every reduction hour in Day~$D$ has its full $L=48$-hour kernel reach contained inside the $T=73$-hour input, requiring no padding.
We choose a 48-hour time window by considering the typical temporal dynamics of pluvial flooding, which can occur rapidly within hours of heavy rainfall, but also considering the need to capture antecedent conditions like soil moisture that can influence flood risk~\cite{bouwens2018towards,upreti2021comparison,james_precipitation_2024}.
A 48-hour kernel allows us to capture both the immediate impact of precipitation and the influence of prior hydrometeorological conditions, providing a comprehensive view of the factors contributing to pluvial flooding.
% (i.e. from $D-2$ to $D$ including the end points) to capture the temporal dynamics of pluvial flooding.
%
% We detail the source and shapes of each input in Table~\ref{tab:inputs}.

\begin{table*}[t]
\centering
\small
\setlength{\tabcolsep}{4pt}
\begin{tabular}{@{}p{0.18\linewidth}p{0.62\linewidth}c@{}}
\toprule
Modality & Source & Shape \\
\midrule
Hydrometeorology & Precipitation (NOAA AORC~\cite{fall_analysis_2023}); soil moisture and snow water equivalent (NWM v3~\cite{cosgrove_noaas_2024}) & $7 \times 7 \times 73 \times 3$ \\
Terrain & HAND~\cite{aws_copernicus_dem}, TWI~\cite{hoylman_30m_2021}, GCN (dry/avg/wet)~\cite{jaafar_gcn250_2019}, POLARIS ($K_{sat}$, clay, $\theta_s$)~\cite{chaney_polaris_2019} & $56 \times 56 \times 8$ \\
Foundation Model & AlphaEarth embeddings~\cite{brown_alphaearth_2025} & $56 \times 56 \times 64$ \\
Exposure & NSI building inventory~\cite{usace_nsi_2022}, NFIP active policies~\cite{fema_nfip_policies_v2} & $56 \times 56 \times 2$ \\
Vulnerability & NSI structural~\cite{usace_nsi_2022}, NFIP policy attributes~\cite{fema_nfip_policies_v2} & $56 \times 56 \times 10$ \\
\bottomrule
\end{tabular}\\[2pt]
\caption{DELUGE's input modalities and tensor shapes $(H \times W \times C)$. For hydrometeorology, the third axis is the $T=73$-step hourly window and the final dimension stacks the three channels; for other modalities the last axis is the feature-channel count.}
    \vspace{-0.15in}
\label{tab:inputs}
\end{table*}

% The three climate variables capture complementary mechanisms relevant to pluvial flooding: APCP is the proximate causal flux; SOIL\_M carries antecedent saturation state; SNEQV accounts for melt contributions. We omit NWM's accumulated underground runoff (UGDROFF)---it represents deep percolation to baseflow, the slowest hydrologic process in the dataset and more relevant to fluvial than pluvial flooding. In ablation it was redundant with SOIL\_M and offered no validation gain.
\subsection{Value Modulator}\label{sec:arch-warp}

Raw hydrometeorology values often span large ranges, and their flood-relevance is nonlinearly distributed across that range. 
For instance, a soil moisture measurement of $0.2~\mathrm{m^3\,m^{-3}}$ may be more relevant to pluvial flood dynamics in one location than in another.
Standard ML pipelines typically apply a global transformation (e.g.\ min-max standardization or a log transform) to the raw values, but we expect the value-relevance relationship to vary across locations, so such a one-size-fits-all transformation may miss this local variability.
This motivates our \emph{Value Modulator}, which learns a per-patch, per-channel transformation of the raw climate values, conditioned on the local terrain and foundation model features which we call a \emph{warp}.

\paragraph{Warp definition.}
For each hydrometeorology channel $v$ and patch $i$, the value modulator produces $K = 16$ width parameters $\{u_{v,i,k}\}_{k=1}^{K}$ and $K = 16$ height parameters $\{h_{v,i,k}\}_{k=1}^{K}$, each softmax-normalized so $\sum_k u_{v,i,k} = \sum_k h_{v,i,k} = 1$.
We choose $K=16$ as a balance between expressivity and simplicity, finding 16 control points sufficient to capture the value-relevance relationship.
These control points define cumulative breakpoints on the input and output axes, so that $X_{v,i,k} \;=\; \sum_{j=1}^{k} u_{v,i,j}$ and $Y_{v,i,k} \;=\; \sum_{j=1}^{k} h_{v,i,j}$,with $X_{v,i,0} = Y_{v,i,0} = 0$ and $X_{v,i,K} = Y_{v,i,K} = 1$. 
The warp $\phi_{v,i} : [0, 1] \to [0, 1]$ is then a monotonic piecewise-linear map satisfying $\phi_{v,i}(X_{v,i,k}) = Y_{v,i,k}$, with slope $m_{v,i,k} = h_{v,i,k} / u_{v,i,k}$ inside the $k$-th segment.
We define the warp, $\phi_{v,i}$, for each $\tilde{x}\in [X_{v,i,k-1},\, X_{v,i,k}]$ as $\phi_{v,i}(\tilde{x}) \;=\; Y_{v,i,k-1} + m_{v,i,k}\,(\tilde{x} - X_{v,i,k-1})$.
%
% \begin{equation}
%     \phi_{v,i}(\tilde{x}) \;=\; Y_{v,i,k-1} + \frac{h_{v,i,k}}{u_{v,i,k}}\,(\tilde{x} - X_{v,i,k-1}) 
% \end{equation}
% Softmax normalization on both $\{u\}$ and $\{h\}$ guarantees monotonicity and a fixed $[0,1] \to [0,1]$ domain/range without further constraints.

\paragraph{Application.}
Each raw input $x_v[t, i]$ is first min-max normalized to $\tilde{x}_v[t, i] \in [0, 1]$ using channel-specific bounds, then the warp is applied to produce the warped value, $\tilde{x}^{\,\prime}_v[t, i] \;=\; \phi_{v,i}\!\left(\tilde{x}_v[t, i]\right)$.
% \begin{equation}
%     \tilde{x}^{\,\prime}_v[t, i] \;=\; \phi_{v,i}\!\left(\tilde{x}_v[t, i]\right)
% \end{equation}
The warp $\phi_{v,i}$ varies per channel and per patch but is \emph{shared across time steps} within the input window, so a single learned warp reshapes every channel $v$ at patch $i$ identically. The warped sequence $\tilde{x}^{\,\prime}_v[t, i]$ then forms the input to the temporal modulator (\S\ref{sec:arch-kernel}).

\paragraph{Per-patch terrain and foundation-model conditioning.}
The $2K \cdot V$ warp parameters per patch are produced by a terrain- and foundation-model-conditioned module \textsc{value\_modulation}, a single convolution layer on the stacked terrain and AEF features followed by a two-layer MLP.
We zero-initialize the weights and biases of this module's final layer, so the softmax over $u_{v,i,k}$ and $h_{v,i,k}$ yields uniform widths and heights at the start of training. The initial warp is therefore the identity map $\phi_{v,i}(\tilde{x}) = \tilde{x}$, and training departs from this identity as each patch's terrain and AEF context pushes its breakpoints apart.
\subsection{Temporal Modulator}
\label{sec:arch-kernel}

% Standard temporal encoders (ConvLSTM, TCN, Transformer) compress this window through opaque operations, foreclosing extraction of the response function the model discovers per cell. We instead use a parametric gamma-kernel module whose learned per-cell parameters admit a direct hydrological interpretation.

\paragraph{Gamma kernels and two-component mixture.}
For each channel $v$ and lag $\tau \in \{0, \ldots, L{-}1\}$ with kernel lookback $L = 48$, a single gamma kernel parameterized by peak lag $p$ and timescale $\theta$ is the discretized, normalized Gamma$(\alpha, \theta)$ density: $k(\tau; p, \theta) \;=\; \frac{\tau^{\alpha - 1}\, e^{-\tau/\theta}}{\sum_{\tau'=0}^{L-1} (\tau')^{\alpha - 1}\, e^{-\tau'/\theta}}$, where $\alpha = \tfrac{p}{\theta} + 1$.
% \begin{equation}
%     k(\tau; p, \theta) \;=\; \frac{\tau^{\alpha - 1}\, e^{-\tau/\theta}}{\sum_{\tau'=0}^{L-1} (\tau')^{\alpha - 1}\, e^{-\tau'/\theta}}, \quad \alpha = \tfrac{p}{\theta} + 1.
% \end{equation}
We reparameterize the shape via $\alpha = p/\theta + 1$ so that the kernel's peak lag is exactly $p$ (the mode of Gamma$(\alpha, \theta)$ lies at $(\alpha - 1)\theta = p$), making the two learned parameters $(p, \theta)$ directly interpretable: $p$ is the peak response time in hours and $\theta$ is the decay timescale in hours. 
%
% In practice, we evaluate $k(\tau)$ as a softmax over $(\alpha - 1)\log(\tau + \varepsilon) - \tau/\theta$ for numerical stability. 
%
The gamma is chosen here as a flexible, two-parameter kernel often used in hydrology to model rainfall-runoff response~\cite{nash_systematic_1959}.
% ; with its single mode and exponential tail, it spans temporal patterns from sharply peaked and fast-decaying to broad and slowly decaying.
% The gamma form arises as the impulse response of a Nash cascade of linear reservoirs~\fix{cite}, a strong inductive bias for rainfall-runoff temporal patterns.

To capture varied temporal patterns, we adopt a per-channel convex mixture of two gammas with the same functional form but different parameters for the precipitation channel.
Here, we parameterize a convex mixture of two gamma, weighted by a learnt mixing weight $w_v$: $k_v(\tau) \;=\; w_v\, k(\tau; p^{\text{1}}_v, \theta^{\text{1}}_v) \;+\; (1 - w_v)\, k(\tau; p^{\text{2}}_v, \theta^{\text{2}}_v)$.
% \begin{equation}
%     k_v(\tau) \;=\; w_v\, k(\tau; p^{\text{1}}_v, \theta^{\text{1}}_v) \;+\; (1 - w_v)\, k(\tau; p^{\text{2}}_v, \theta^{\text{2}}_v),
% \end{equation}
%
For soil moisture and snow water equivalent, we employ a single gamma.
%
% with the ordering $p^{\text{s}}_v = p^{\text{f}}_v + (L - p^{\text{f}}_v)\,\rho_v$ enforced, where $\rho_v \in (0, 1)$ is a sigmoid-bounded learned offset; this guarantees $p^{\text{s}}_v > p^{\text{f}}_v$ and prevents the fast and slow modes from being swappable across training runs. 
% All other kernel parameters are sigmoid-bounded: $p_v \in [0.5, L]$, $\theta \in [0.5, L]$~hours, $w \in (0, 1)$. 
% 
% We bound $p_v$ below with 0.5 to avoid numerical instability.
%
In all, each channel's temporal response kernel is parameterized by five (precipitation) or two (rest) parameters, totaling  kernel parameters per patch.

\paragraph{Per-patch terrain conditioning.}
The \textsc{temporal\_modulation} module is structurally analogous to \textsc{value\_modulation}, producing the 9 kernel parameters per patch from the stacked terrain and AEF features.
% The kernel parameters are produced per cell by a AEF\&terrain-conditioned module: a single $3 \times 3$ ConvBlock ($C_t \to 64$) over the stacked terrain and AEF features $\mathbf{t}$ followed by global average pooling and a two-layer MLP to produce the 15 parameters per sample.
%
% The ConvBlock preserves the terrain texture (slope variance, imperviousness patchiness) that pure pooling would discard. 
%
The final linear layer is zero-initialized in weight, with bias chosen so each cell decodes the same hydrologically reasonable kernel at step 0 ($p^{\text{1}} \approx 3$h, $p^{\text{2}} \approx 24$h, $\theta^{\text{1}} \approx 3$h, $\theta^{\text{2}} \approx 12$h, $w \approx 0.5$); training then pushes per-patch parameters apart based on each patch's characteristics.

\paragraph{Causal convolution}
Let $i$ be one of the pixels within the $7 \times 7$ patch, and let $x_v[t, i]$ denote the value of hydrometeorology channel $v$ at time $t$ and pixel $i$ within the $T$-hour input window. The kernel $k_v$ is then convolved causally with the input sequence to produce the response at pixel $i$: $r_v[t, i] \;=\; \sum_{\tau=0}^{L-1} k_v(\tau)\, x_v[t - \tau,\, i]$.
The kernel is shared across all pixels within the patch.
\paragraph{Temporal reduction}
The temporal axis is collapsed by selecting one time step $t'$ per pixel, the moment of peak accumulated precipitation \emph{response} within Day~$D$: $t'(i) = \operatorname*{argmax}_{t \in \text{Day D}} r_{\text{precip}}[t, i]$. 
We reduce this via, $r_{\text{reduced}}[v, i] = r_v[t'(i), i]$.
All channels are read at the same $t'$, yielding a temporally-coherent Day-$D$ snapshot.
We expect this to summarize the hydrometeorological state at the moment of peak precipitation impact, appropriately weighing the lookback hours according to the learned kernel.
Finally, a linear $1 \times 1$ convolution lifts the per-pixel $V$-dimensional reduced response to a 32D feature space for fusion with the other modalities.
% \begin{equation}
%     \mathbf{h} \;=\; \mathrm{Dropout}\!\left(\mathrm{LeakyReLU}(\mathbf{W}\, r_{\text{red}} + \mathbf{b})\right), \quad \mathbf{h} \in \mathbb{R}^{B \times 16 \times H \times W}.
% \end{equation}
% We use a linear projection rather than a nonlinear bottleneck: nonlinear heads back-pressure the kernel parameters during training (Section~\ref{sec:arch-design}), eroding the interpretability of the learned $(p, \theta)$ maps. The kernel encoder itself uses $\sim$80 parameters; the dominant cost is the terrain conditioner at $\sim$15K parameters---roughly one-quarter of an equivalent ConvLSTM encoder.

\vspace{-3pt}
\subsection{Other Components}
\paragraph{Modality Encoders.}

The remaining three modalities are processed by lightweight convolutional encoders, all operating on the same \tsim$8\times8$\,km patch.
The NSI building-inventory and structural fields are native to the $56 \times 56$ grid, while the NFIP active-policy and policy-attribute fields are native to the $7 \times 7$ grid and are bilinearly upsampled to $56 \times 56$, so the \emph{Terrain-Hazard}, \emph{Exposure}, and \emph{Vulnerability} encoders all operate at a common $56 \times 56$ resolution.
Each encoder maps its input through strided convolutional blocks that downsample to $7 \times 7$, producing 16, 8, and 8 channels respectively.
All encoders use $3 \times 3$ convolutions, leaky-ReLU activations, and batch normalization~\cite{ioffe_batch_2015}.

\paragraph{Cross-Modal Fusion.}

At each of the $7 \times 7$ spatial positions, the four modality feature vectors form a set of four tokens.
Each token is linearly projected to $d_{\text{model}} = 32$ by a $1 \times 1$ convolution and summed with a learnable modality-type embedding that identifies its source modality.
We fuse the tokens with a single Transformer encoder layer~\cite{vaswani_attention_2017}, applying 4-head self-attention and a position-wise feed-forward network, each wrapped in a residual connection and layer normalization, so that every modality attends to the others before fusion.
The fused representation $f$ is read out at the center pixel to align with the per-pixel prediction target.

\paragraph{Classifier.}
\label{sec:arch-classifier}

The center-pixel feature vector $f \in \mathbb{R}^{d_{\text{model}}}$ is mapped to a claim probability by an MLP classification head with a single hidden layer.
The hidden layer applies layer normalization and a GELU nonlinearity~\cite{hendrycks_gaussian_2016}, and a final linear layer with a sigmoid activation produces the per-pixel probability $\hat{y} \in [0,1]$.

\subsection{Training}
\label{sec:arch-training}

\paragraph{Spatial Holdout Split.}
To prevent spatial leakage, we partition the 100 grid cells (\S\ref{ssec:study_region}) into train and test sets at the 75\,km cell level rather than at the pixel level.
We use a spatially blocked split of grid cells, so the model is evaluated on held-out geographic regions rather than on held-out samples from the same regions.
Because our 100 grid cells span the diverse hydroclimatic regions of CONUS, across which flood dynamics are known to vary significantly~\cite{brunner_spatial_2020}, evaluating on spatially held-out cells tests whether DELUGE generalizes across this wide range of flood dynamics rather than to new samples from cells it has already seen.
\paragraph{Loss.} We use focal loss~\cite{lin_focal_2020} with $\alpha = 0.5$, $\gamma = 2$, weighted per sample by our claim polygon area based confidence score $c$ (\S\ref{ssec:dataset_construction}).
This choice is primarily motivated by the extreme class imbalance and to de-emphasize the easy negative samples, which are abundant in the dataset and can overwhelm the learning signal from the more informative positive samples.
%
% We empirically found focal loss to outperform binary cross-entropy, likely because it better handles the extreme class imbalance and suppresses easy minority class samples.

In addition to the main loss on the predicted flood probability, we add a smoothness penalty on the Value Modulator's parameters (\S\ref{sec:arch-warp}):
$\mathcal{L}_{\text{smooth}}=\sum_{v} \frac{1}{K-1} \sum_{k=1}^{K-1} \left[(u_{v,k+1} - u_{v,k})^2 + (h_{v,k+1} - h_{v,k})^2\right]$.
% \begin{equation}
%     \mathcal{L}_{\text{smooth}}=\sum_{v} \frac{1}{K-1} \sum_{k=1}^{K-1} \left[(u_{v,k+1} - u_{v,k})^2 + (h_{v,k+1} - h_{v,k})^2\right]
% \end{equation}
%
The penalty encourages adjacent widths and adjacent heights to vary smoothly across segments, biasing the learned warp toward a smooth function.
% The softmax constraints already enforce monotonicity and the $[0, 1] \to [0, 1]$ boundary conditions; the smoothness term is the additional inductive bias that keeps the learned warps human-readable.
% 
In all, we optimize for the weighted sum of the focal loss and the smoothness penalty:
$\mathcal{L} \;=\; \mathcal{L}_{\text{focal}} + \lambda\cdot\mathcal{L}_{\text{smooth}}$ with $\lambda = 0.1$.
% \begin{equation}
% \mathcal{L} \;=\; \mathcal{L}_{\text{focal}} + \lambda\cdot\mathcal{L}_{\text{smooth}}, \quad \lambda = 0.1
% \end{equation}
We apply the AdamW optimizer~\cite{loshchilov_decoupled_2019} with a cosine learning-rate schedule and linear warmup.

\section{Results}
\begin{figure*}[!ht]
\centering
\setlength{\tabcolsep}{1pt}
\renewcommand{\arraystretch}{0.6}
\begin{tabular}{c@{\hspace{1pt}}ccccccc}
\rotatebox{90}{\,\textbf{NFIP Claims}} &
\includegraphics[width=0.135\textwidth]{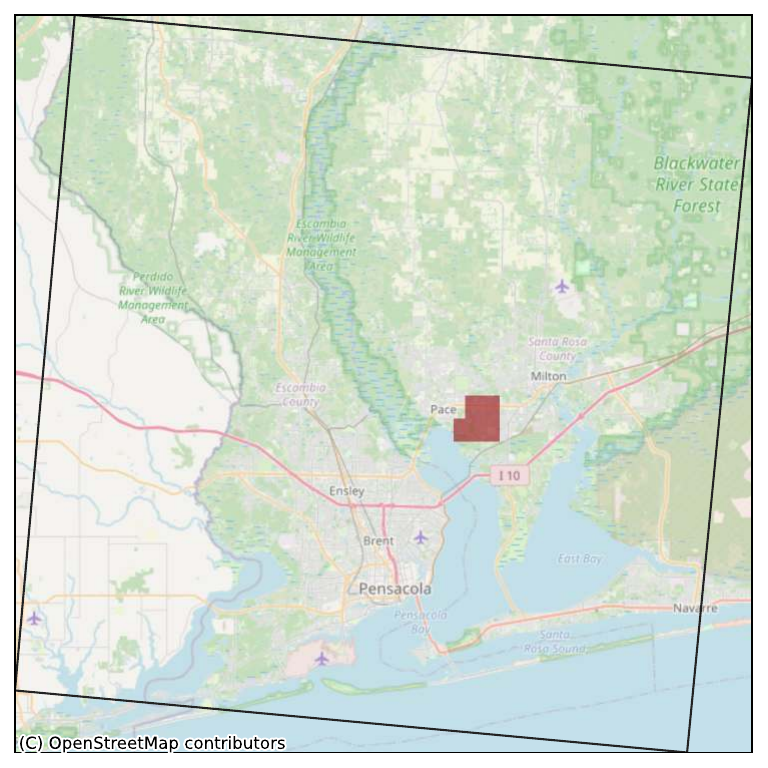} &
\includegraphics[width=0.135\textwidth]{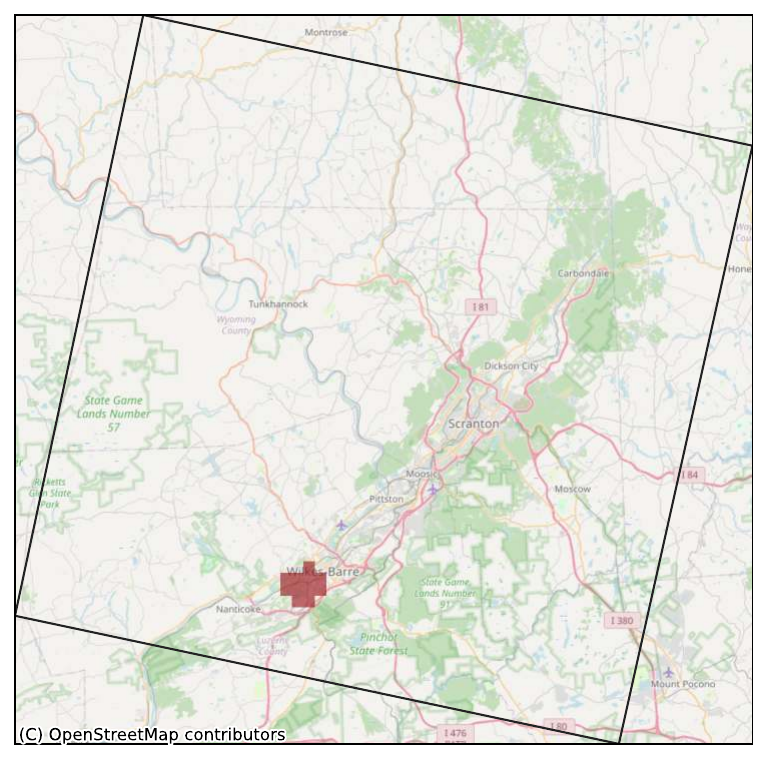} &
\includegraphics[width=0.135\textwidth]{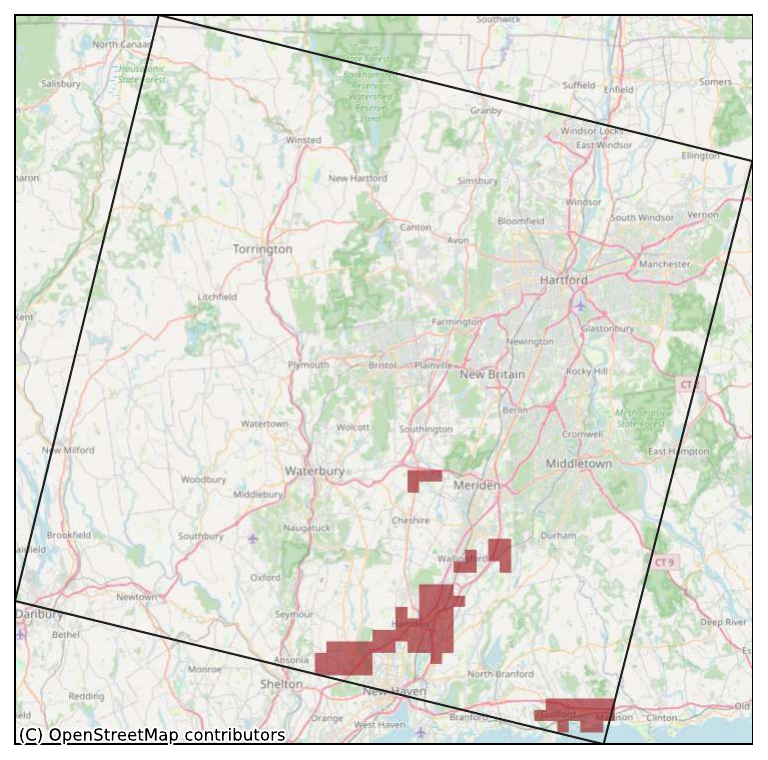} &
\includegraphics[width=0.135\textwidth]{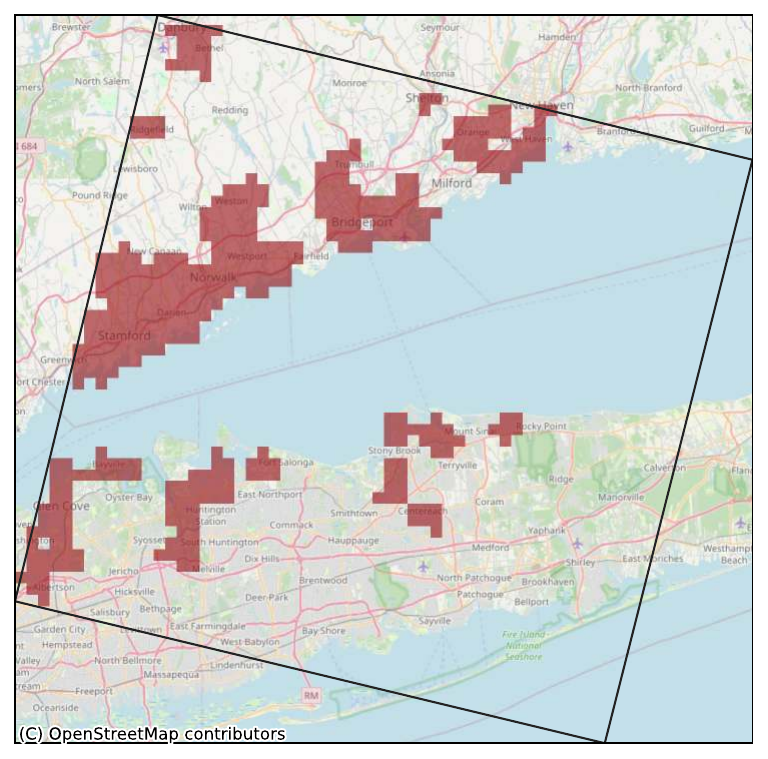}&
\includegraphics[width=0.135\textwidth]{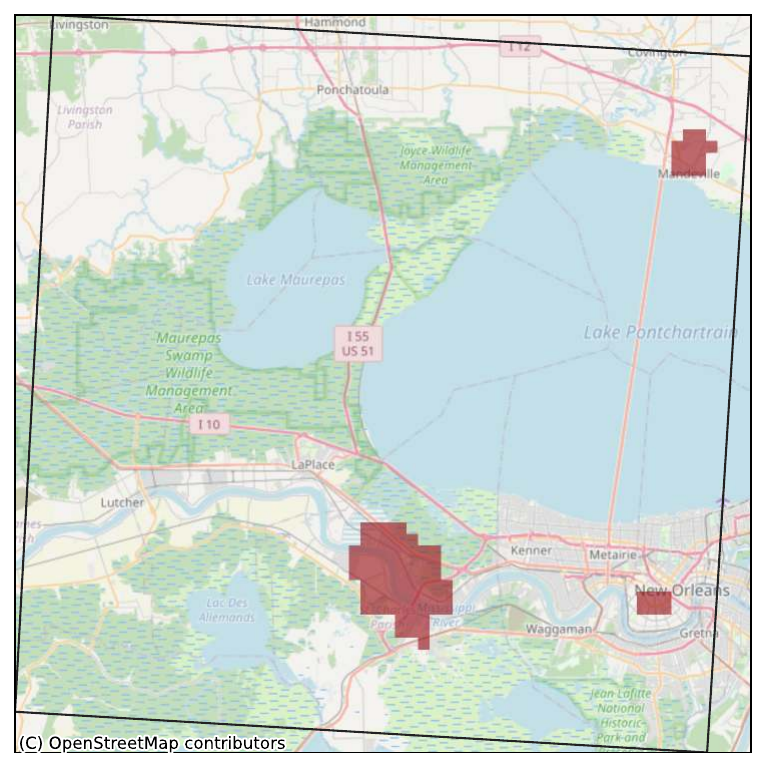} &
\includegraphics[width=0.135\textwidth]{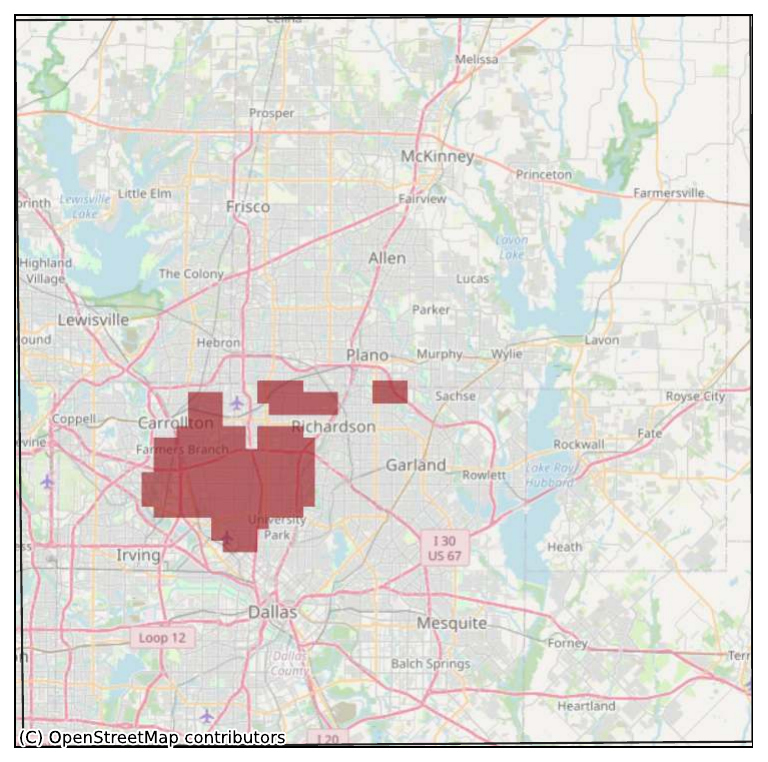} &
\includegraphics[width=0.135\textwidth]{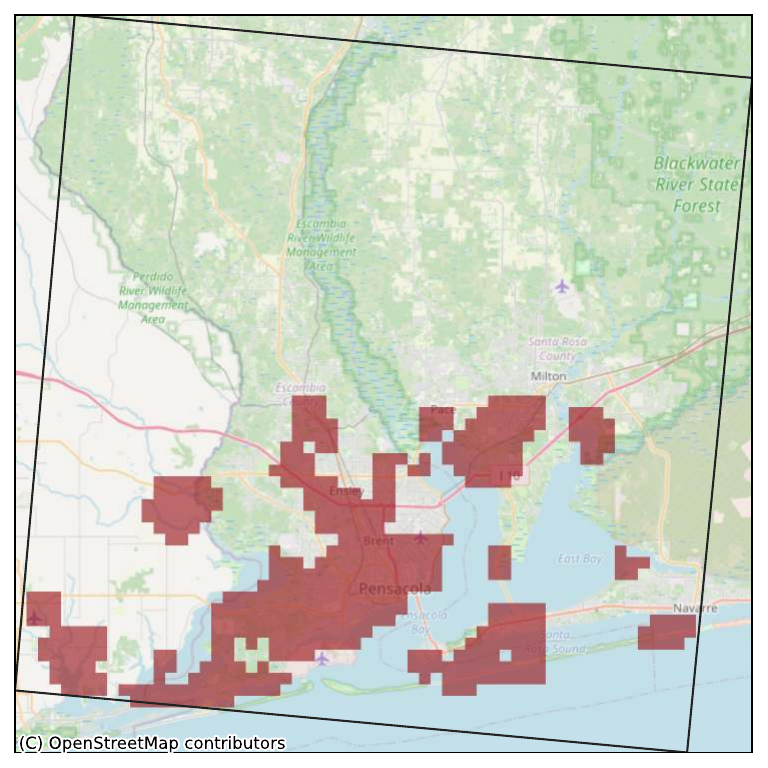} \\
\rotatebox{90}{\,\textbf{Predicted}} &
\includegraphics[width=0.135\textwidth]{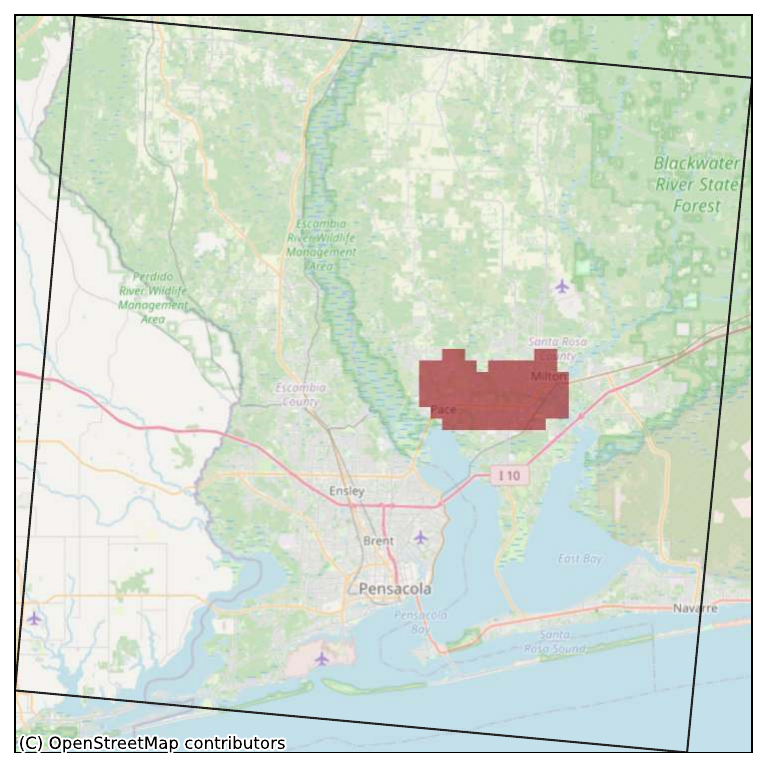} &
\includegraphics[width=0.135\textwidth]{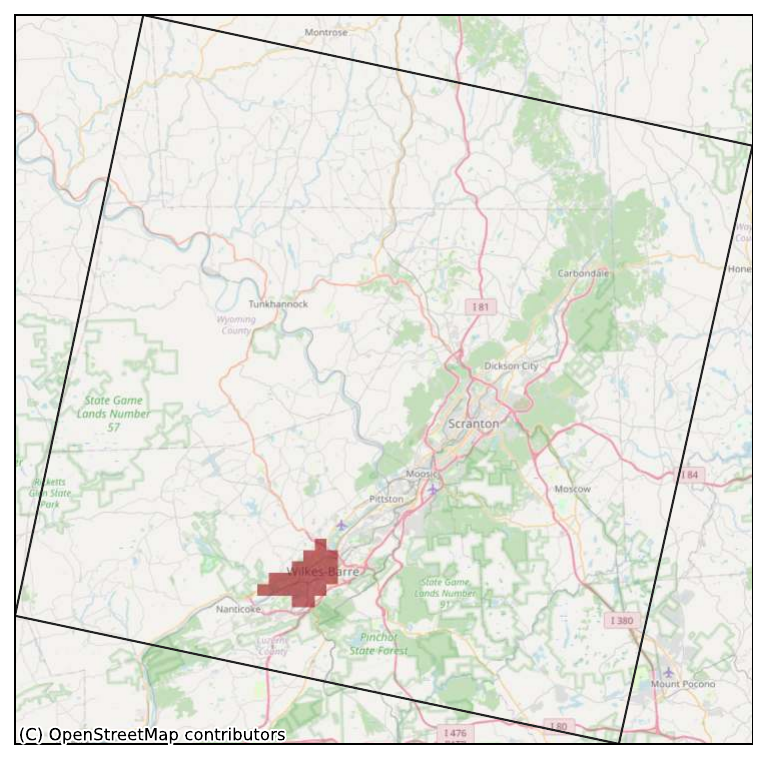} &
\includegraphics[width=0.135\textwidth]{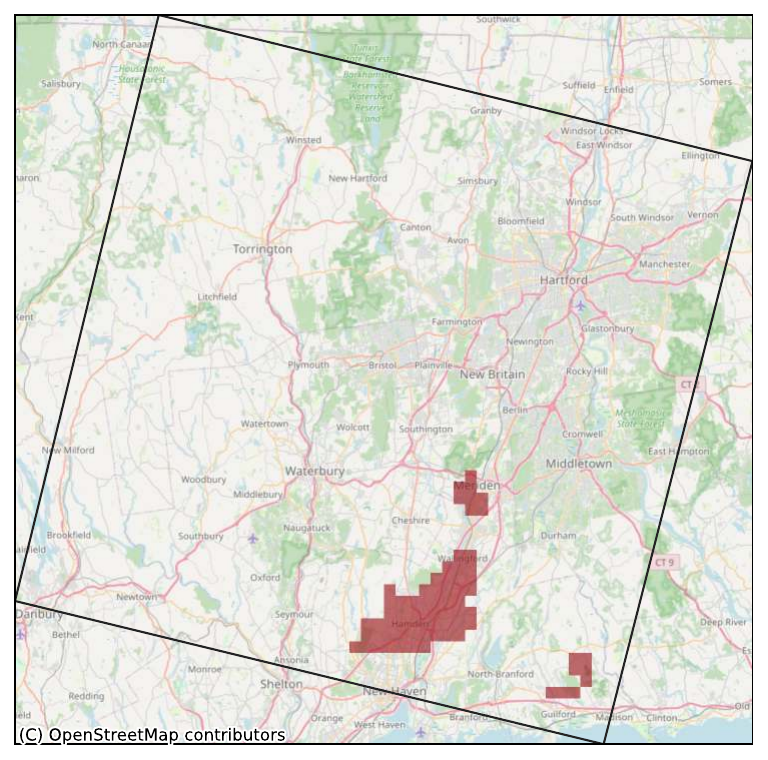} &
\includegraphics[width=0.135\textwidth]{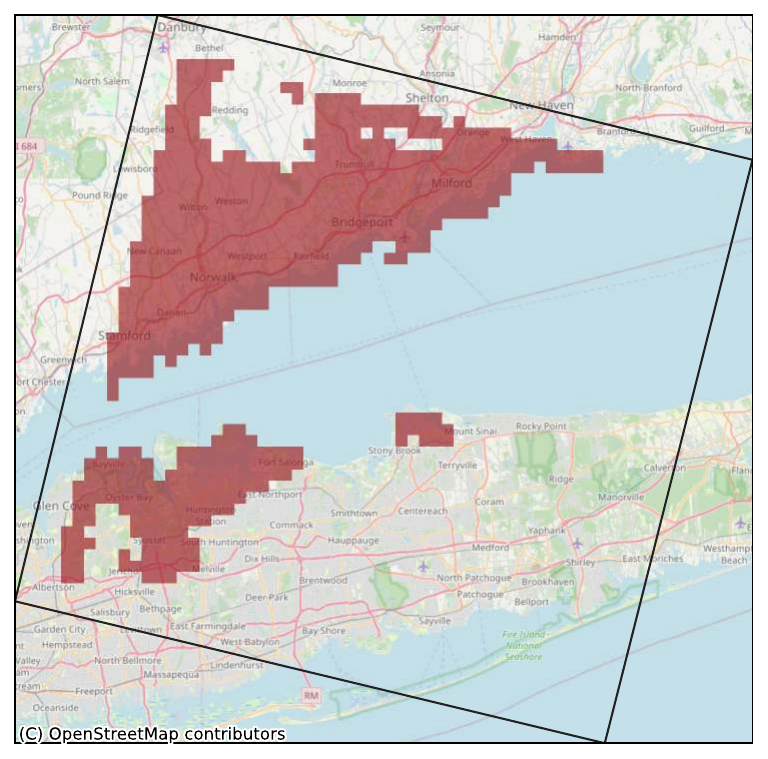} &
\includegraphics[width=0.135\textwidth]{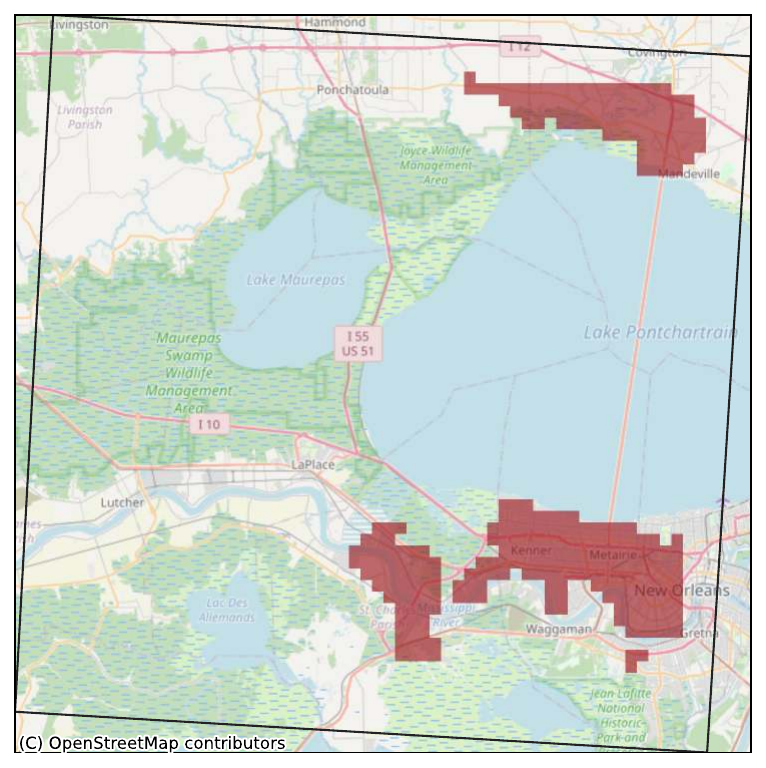} &
\includegraphics[width=0.135\textwidth]{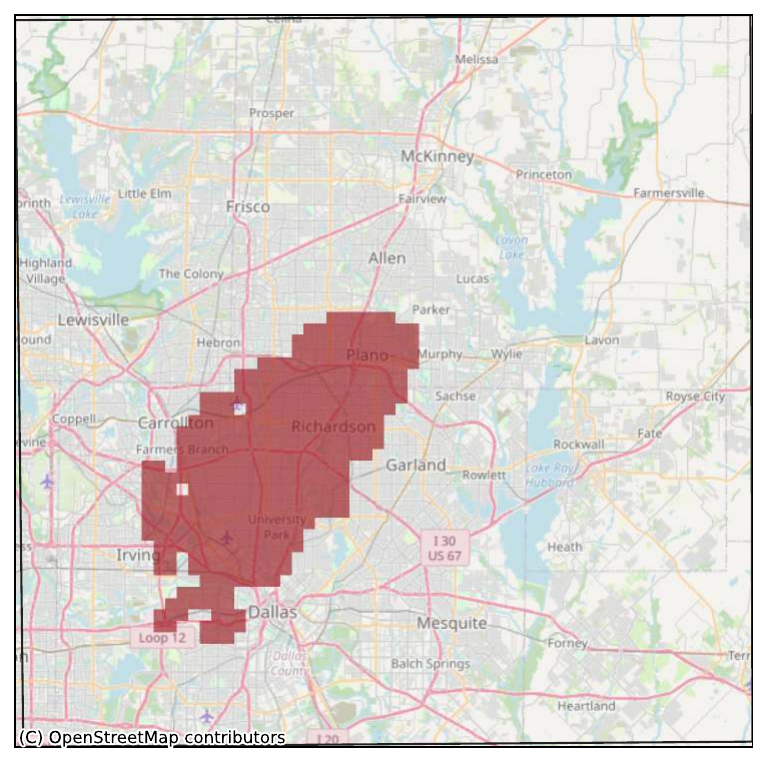} &
\includegraphics[width=0.135\textwidth]{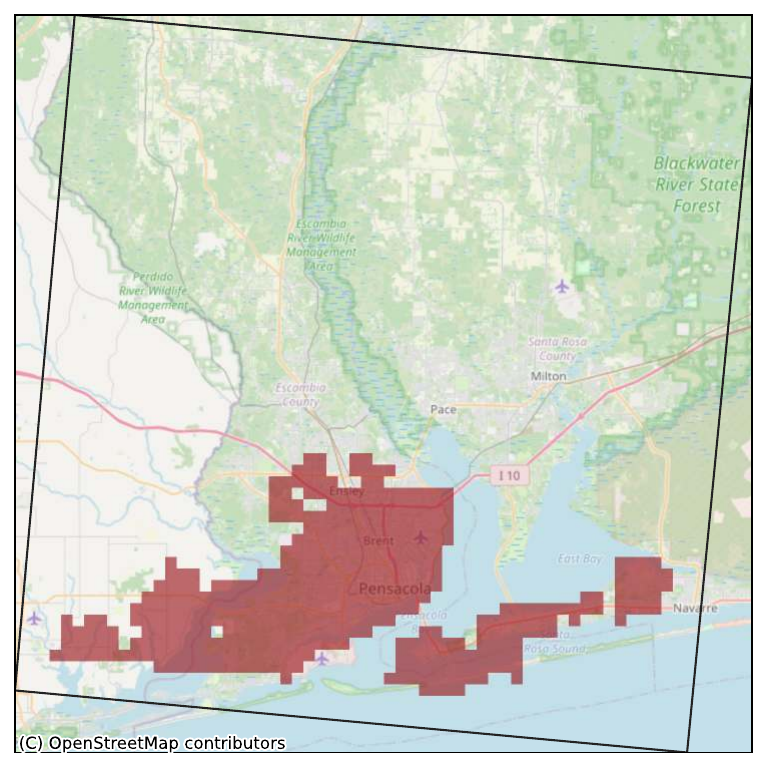} \\
 % & (a) & (b) & (c) & (d) & (e) & (f) & (g) \\
\end{tabular}
\caption{Pluvial NFIP claims (top) and DELUGE predictions (bottom) on seven held-out test days. Red colored cells corresponds to pixels with flood claims (top) and pixel where DELUGE predicted pluvial NFIP flood claims (bottom).}
    \vspace{-0.15in}
\label{fig:obs-pred}
\end{figure*}
To evaluate the performance of DELUGE, we compare it against tree-based baselines, specifically Random Forest, XGBoost~\cite{chen_xgboost_2016}, and LightGBM~\cite{ke_lightgbm_2017}.
We focus on tree-based baselines for two reasons.
First, the existing ML literature on flood-damage prediction overwhelmingly relies on tabular models, with Random Forest the most commonly reported baseline~\cite{yang_predicting_2022,alipour_leveraging_2020,collins_predicting_2022,garcia_reconstructing_2025,liao_fast_2023}; we therefore include Random Forest alongside XGBoost and LightGBM as stronger tuned variants of the same family.
Second, while graph neural network (GNN) approaches have been proposed for flood prediction~\cite{sarkar_hydrogat_2025,kazadi_floodgnn-gru_2024,taghizadeh_interpretable_2024}, they have been developed primarily for fluvial dynamics, predicting inundation depth, and demonstrated on small catchment-scale domains, making them impractical for a CONUS-wide pluvial setting.
\paragraph{Evaluation Metrics.}Given the severe class imbalance of our problem, we use the Precision-Recall AUC (PR-AUC) as our evaluation metric.
In addition, we develop a dollar-weighted version of PR-AUC to account for varied range in reported flood damages.
To this end, we associate each positive sample with the damage density where the claim damage is equally distributed across the pixels in the claim's polygon footprint.
Then, integrating under a precision-recall curve where the recall is defined as the percentage of total claim dollars correctly predicted as flood claims, with precision as its standard definition, we obtain a \textbf{dollar-weighted PR-AUC metric that captures the model's ability to identify the claims weighted by claimed damages.}
Under this metric, we evaluate the performance of our model in terms of its ability to capture the most costly flood claims, which is of interest to insurance companies and other stakeholders.

\vspace{-5pt}
\subsection{Comparison against Baselines}Table~\ref{tab:comparison} presents the headline performance comparison between DELUGE and baseline models.
%
% tree based models dont really take in time series
Since tree-based models do not natively handle time-series or spatial rasters, we use the same set of features as introduced in \S\ref{ssec:arch-inputs}, but with some modifications to make them compatible with tree-based models.
For the hydrometeorological time-series, we extract the daily sum and daily max for each of the hydrometeorological features as often done in literature~\cite{yang_predicting_2022, alipour_leveraging_2020} and the center pixel value and the spatial mean for all other features in the given patch. 
We tuned the tree-based baselines (RF, XGBoost, LightGBM) by randomized search over 50 configurations.

\textbf{DELUGE outperforms all three baselines on both metrics}, with LightGBM the strongest tree baseline (Table~\ref{tab:comparison}).
% , LightGBM close behind, and Random Forest lagging substantially 
%
% Because each comparison shares the same splits, $\Delta\%$ is the mean per-seed difference rather than a ratio of marginal means, and DELUGE leads every baseline on this paired basis, so the overlapping absolute error bars reflect shared seed-to-seed variance rather than ambiguity in the ranking.
Because every method is evaluated on the same splits, $\Delta\%$ is a mean per-seed difference rather than a ratio of marginal means.
DELUGE leads every baseline on this paired basis, so the overlapping absolute-error bars reflect shared seed-to-seed variance, not ambiguity in the ranking.
The relative gap widens on the dollar-weighted metric, with the tree baselines trailing DELUGE by \tsim$9$ to $30\%$ on \$-Weighted PR-AUC versus \tsim$6$ to $27\%$ on PR-AUC.
This pattern suggests that \textbf{DELUGE's architecture helps capture the high-cost claims that tabular baselines under-predict}, the regime most relevant to insurance and emergency-management stakeholders.
Figure~\ref{fig:obs-pred} shows observed claims alongside DELUGE predictions drawn from the held-out spatial split.

\begin{table}[]
\small
\centering
\setlength{\tabcolsep}{4pt}
\begin{tabular}{l r r r r}
\toprule
\textbf{Model} & \textbf{PR-AUC} & \textbf{$\Delta\%$} & \textbf{\$-W PR-AUC} & \textbf{$\Delta\%$} \\
\midrule
\textbf{DELUGE (Ours)} & $\mathbf{0.243_{\pm 0.007}}$ & --      & $\mathbf{0.594_{\pm 0.049}}$ & --      \\
Random Forest          & $0.177_{\pm 0.014}$          & $-27\%$ & $0.413_{\pm 0.016}$          & $-30\%$ \\
XGBoost~\cite{chen_xgboost_2016}                & $0.219_{\pm 0.006}$          & $-10\%$  & $0.511_{\pm 0.018}$          & $-14\%$ \\
LightGBM~\cite{ke_lightgbm_2017}               & $0.228_{\pm 0.015}$          & $-6\%$ & $0.538_{\pm 0.030}$          & $-9\%$ \\
\bottomrule
\end{tabular}
\caption{\textbf{Comparison of DELUGE against tree-based baselines.} PR-AUC measures binary damage-occurrence prediction, and \$-Weighted PR-AUC weights predictions by claim amount. At a \tsim$0.25\%$ prevalence, DELUGE shows a \tsim$100$x improvement over chance performance. Scores are reported over three matched spatial splits, and $\Delta\%$ is each baseline's mean per-seed relative change from DELUGE. }
\label{tab:comparison}
\end{table}
\paragraph{Operating-Point Analysis.}
To probe how this dollar concentration manifests at fixed alert budgets, we report top-$K$ recall at several values of $K$ (Table~\ref{tab:topk}), separating occurrence recall (R@K) from dollar recall (\$R@K).
At every $K$, \$R@K sharply exceeds R@K.
At $K=0.1\%$ DELUGE captures only $19\%$ of pluvial claim events but $62\%$ of total damage dollars and at $K=1\%$ the gap remains ($52\%$ events vs.\ $90\%$ dollars).
DELUGE also leads both XGBoost and LightGBM at every $K$ on both metrics, with the largest gap at the tightest, most operationally relevant budgets.
\textbf{DELUGE's top-ranked predictions are therefore concentrated in the high-cost regime}, the operationally meaningful slice of the test set for insurance and emergency-management workflows.

\vspace{5pt}
\begin{table}[h]
\small
\centering
\setlength{\tabcolsep}{4pt}
\begin{tabular}{l rr rr rr}
\toprule
& \multicolumn{2}{c}{\textbf{DELUGE}} & \multicolumn{2}{c}{XGBoost} & \multicolumn{2}{c}{LightGBM} \\
\cmidrule(lr){2-3}\cmidrule(lr){4-5}\cmidrule(lr){6-7}
$\boldsymbol{K}$ & R@K & \$R@K & R@K & \$R@K & R@K & \$R@K \\
\midrule
$0.1\%$ & $\mathbf{0.193}$ & $\mathbf{0.617}$ & 0.173 & 0.609 & 0.180 & 0.557 \\
$0.2\%$ & $\mathbf{0.279}$ & $\mathbf{0.728}$ & 0.257 & 0.701 & 0.260 & 0.668 \\
$0.5\%$ & $\mathbf{0.412}$ & $\mathbf{0.832}$ & 0.388 & 0.813 & 0.391 & 0.803 \\
$1\%$   & $\mathbf{0.521}$ & $\mathbf{0.897}$ & 0.495 & 0.880 & 0.500 & 0.872 \\
\bottomrule
\end{tabular}
\caption{Top-$K$ recall. R@K is occurrence recall at $K$ (fraction of positive pixel-days captured) and \$R@K is dollar recall at $K$ (fraction of total claim dollars captured). $K$ given as a percentile of the held-out pixel-day test set.}
% \fix{adding error bars}}
\label{tab:topk}
\end{table}
\vspace{-5pt}
\subsection{Ablation Studies}
We ablate DELUGE along two axes (Table~\ref{tab:ablations}). \emph{Module ablations} swap out our architectural contributions and \emph{feature ablations} drop one input modality at a time while leaving the architecture intact.
\paragraph{Modulator Ablations.} We replace our modulator-based encoder with a standard ConvLSTM and a Transformer, common architectures for spatiotemporal data, applied to the hydrometeorological time-series.
DELUGE performs on par with both alternatives on mean, with the Transformer marginally ahead on PR-AUC and DELUGE marginally ahead on \$-Weighted PR-AUC, while ConvLSTM trails slightly on both.
\textbf{Our modulator design therefore retains predictive accuracy comparable to stronger sequence-model alternatives while adding \emph{interpretability-by-design}.}
We then progressively ablate the modulators' conditioning, from frozen at initialization (Uniform), to learnable but globally shared, to per-patch conditioned on a single channel (AEF-only or Terrain-only).
Each step closes part of the gap to full DELUGE on both metrics, confirming that \textbf{both learning the modulator parameters and conditioning them per patch matter}.
AEF-only and Terrain-only perform nearly identically on PR-AUC, with AEF-only modestly stronger on \$-Weighted PR-AUC, suggesting the two conditioning channels carry largely overlapping information about local hydrology.
\textbf{This demonstrates the utility of AlphaEarth embeddings for pluvial flood prediction.} As a conditioning signal they perform on par with a curated suite of hydrological terrain descriptors and well above unconditioned modulators, with no hydrological feature engineering required.

\vspace{-3pt}
\paragraph{Feature Ablations.} The single largest drop comes from removing the hydrometeorological hazard inputs (precipitation, soil moisture, and snow water equivalent), which collapses the model and \textbf{confirms hydrometeorology as the dominant driver of DELUGE's prediction}.
Among the remaining modalities, Exposure shows the largest gap, reflecting its role in identifying which structures sit in the path of damaging precipitation.
%
% Interestingly, removing terrain features together with the AlphaEarth embedding (the modulators' conditioning input) costs a comparable $-$10\% / $-$11\%, but notably smaller than the Uniform Modulators gap above, suggesting that terrain and AlphaEarth contribute more strongly through the modulators than as direct fused features.
%
Vulnerability shows the smallest gap, suggesting structural attributes carry less independent signal, perhaps as they correlate with AlphaEarth~\cite{bell_earth_2026}.

\begin{table}[ht]
\centering
\small
\setlength{\tabcolsep}{3pt}
\begin{tabular}{l r r r r}
\toprule
\textbf{Configuration} & \textbf{PR-AUC} & \textbf{$\Delta\%$} & \textbf{\$-W PR-AUC} & \textbf{$\Delta\%$} \\
\midrule
\textbf{DELUGE (Full)} & $\mathbf{0.243_{\pm 0.007}}$ & -- & $\mathbf{0.594_{\pm 0.049}}$ & -- \\
\midrule
\multicolumn{5}{l}{\emph{Module ablations}} \\
\, Uniform Modulators     & $0.196_{\pm 0.014}$ & $-$19\% & $0.414_{\pm 0.050}$ & $-$30\%  \\
\, Globally-shared Modulators & $0.217_{\pm 0.011}$ & $-$11\% & $0.528_{\pm 0.053}$ & $-$11\% \\
\, AEF-only Modulators & $0.234_{\pm 0.016}$ & $-$4\% & $0.572_{\pm 0.081}$ & $-$4\% \\
\, Terrain-only Modulators & $0.235_{\pm 0.012}$ & $-$3\% & $0.558_{\pm 0.073}$ & $-$6\% \\
\addlinespace[0.3em]
\hdashline[2pt/2pt]
\addlinespace[0.3em]
\, ConvLSTM Hydromet.   & $0.234_{\pm 0.014}$ & $-$4\% & $0.543_{\pm 0.089}$ & $-$9\% \\
\, Transformer Hydromet.   & $0.250_{\pm 0.029}$ & $+$3\% & $0.574_{\pm 0.119}$ & $-$4\% \\
\midrule
\multicolumn{5}{l}{\emph{Feature ablations}} \\
\, Exposure       & $0.217_{\pm 0.017}$ & $-$11\% & $0.476_{\pm 0.071}$ & $-$20\% \\
\, Vulnerability  & $0.229_{\pm 0.012}$ & $-$6\% & $0.526_{\pm 0.056}$ & $-$11\% \\
\, Terrain Hazard & $0.218_{\pm 0.007}$ & $-$11\% & $0.496_{\pm 0.078}$ & $-$17\% \\
\, Hydromet. Hazard & $0.020_{\pm 0.002}$ & $-$92\% & $0.038_{\pm 0.008}$ & $-$94\% \\
\bottomrule
\end{tabular}
\caption{\textbf{Module and feature ablations of DELUGE.} Each row removes or replaces the listed component from the full model. 
Module ablations swap an architectural component while feature ablations drop one input modality while leaving the architecture intact. 
% The Terrain Hazard ablation removes terrain features together with the AlphaEarth embedding, since they jointly condition the modulators. 
}
\label{tab:ablations}
\end{table}

\begin{figure*}[!ht]
\centering
\setlength{\tabcolsep}{1pt}
\renewcommand{\arraystretch}{0.4}
\begin{tabular}{ccccc}
% \rotatebox{90}{\,\textbf{Basemap}} &
% \includegraphics[width=0.18\textwidth]{figs/cell_dynamics_val_cell_1331_basemap.pdf} &
% \includegraphics[width=0.18\textwidth]{figs/cell_dynamics_val_cell_1486_basemap.pdf} &
% \includegraphics[width=0.18\textwidth]{figs/cell_dynamics_val_cell_1136_basemap.pdf} &
% \includegraphics[width=0.18\textwidth]{figs/cell_dynamics_val_cell_1544_basemap.pdf} &
% \includegraphics[width=0.18\textwidth]{figs/cell_dynamics_val_cell_2168_basemap.pdf} \\
% \rotatebox{90}{\,\textbf{Dynamics}} &
\includegraphics[width=0.19\textwidth]{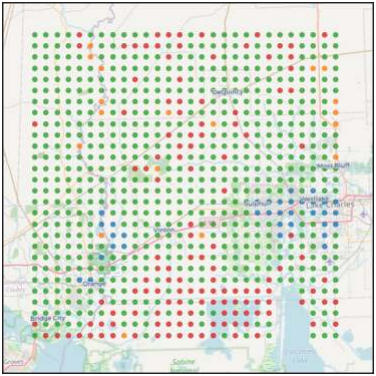} &
\includegraphics[width=0.19\textwidth]{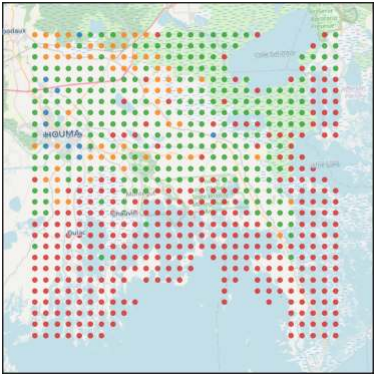} &
\includegraphics[width=0.19\textwidth]{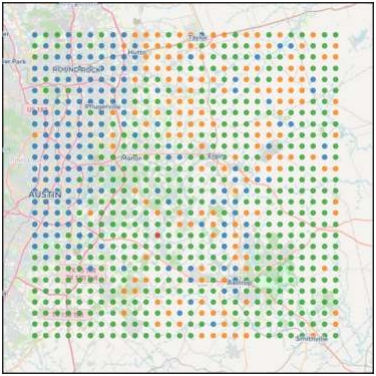} &
\includegraphics[width=0.19\textwidth]{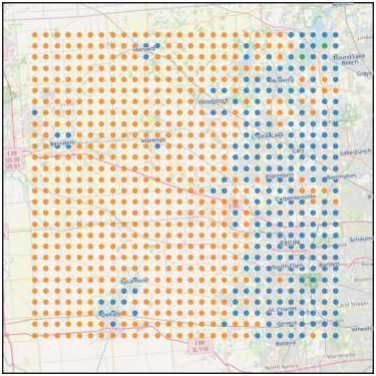} &
\includegraphics[width=0.19\textwidth]{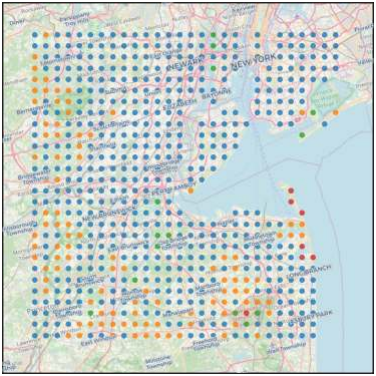} \\
 % & (a) & (b) & (c) & (d) & (e) \\
\end{tabular}
% \caption{Basemap (top) and learned modulator dynamics clustering result (bottom) for five representative held-out cells (\S\ref{ssec:modulator_behaviors}). Colors follows the same scheme as Fig.~\ref{fig:modulator-clusters} for consistency. Note that we take every other pixel for simplicity and clarity.}
\caption{Learned modulator dynamics clustering result for five representative held-out cells (\S\ref{ssec:modulator_behaviors}). Colors follows the same scheme as Fig.~\ref{fig:modulator-clusters} for consistency.}
    \vspace{-0.15in}
\label{fig:cell-dynamics}
\end{figure*}
\vspace{-3pt}
\subsection{Verifying Learned Behavior of the Modulators}\label{ssec:modulator_behaviors}
Explainability and physical fidelity are growing concerns in Geospatial AI~\cite{hsu_explainable_2023,roussel2025introducing,suri_trusting_nodate} and flood modeling~\cite{taghizadeh_interpretable_2024}.
While a substantial body of work examines what geospatial embeddings encode about physical space~\cite{benavides2026earth,rahman_physically_2026}, comparatively little addresses how downstream models should consume these representations in a physically verifiable manner.
DELUGE's modulators are designed to this end. Because their parameters constitute the model's predictive computation directly, the interpretability they afford is architectural rather than post-hoc, obtained by inspecting the learned parameters themselves rather than by explaining individual predictions.

\begin{figure}[]
\centering
\includegraphics[width=\columnwidth]{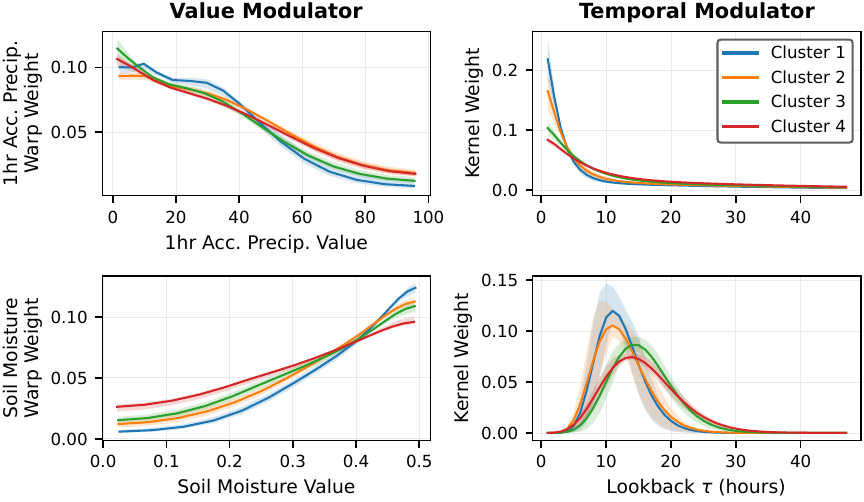}
\caption{Clusters identified in modulator-parameter space ($K=4$) and their corresponding modulator dynamics.}
% \vspace{-10pt}
\label{fig:modulator-clusters}
\end{figure}

The modulators act on the hydrometeorology branch, which our ablations also identify as the dominant predictive driver (Table~\ref{tab:ablations}), so its learned response is the most consequential to interpret.
For each patch, we summarize the Value Modulator's warp $\phi_{v,i}(\tilde{x})$ and Temporal Modulator's response kernel $k_v(\tau)$ by their CDFs.
Here, we omit snow water equivalent and focus on precipitation and soil moisture, the channels most directly relevant to pluvial flooding.
% We compute these for the two channels relevant to pluvial flooding --- precipitation and soil moisture --- omitting snow water equivalent, yielding four CDF blocks per pixel that we standardize and concatenate.
%
To cluster, we apply PCA on the CDFs down to two components (\tsim$75\%$ of variance) followed by K-means. 
A silhouette sweep determined $K=4$.
To illustrate the learned modulator behavior within each cluster, Figure~\ref{fig:cell-dynamics} shows representative held-out grid cells with their modulator dynamics clustering assignments, and Figure~\ref{fig:modulator-clusters} shows the corresponding modulator dynamics for each cluster.

First, examining the Temporal Modulator's kernel (Fig.~\ref{fig:modulator-clusters}) for 1hr acc. precip. (top) and soil moisture (bottom), we find that \textbf{DELUGE learns to rely on precipitation for the immediate risk and soil moisture to capture antecedent conditions}, with the latter kernels peaking in the \tsim$7-15$ hour range in contrast with the precipitation kernels that peak at $\tau\approx0$ and decay rapidly.
Notice in Fig~\ref{fig:modulator-clusters} that within \emph{every} cluster the two kernels are cleanly staggered, with the precipitation kernel concentrated at the most recent hour and the soil-moisture kernel peaking several hours later, an explicit hand-off from instantaneous precipitation forcing to antecedent wetness state that DELUGE recovers without a hydrological prior.
Next, with a further qualitative inspection of the clustering results, we derive the characters of the four identified clusters.

%
% \begin{enumerate}[nosep, align=left, leftmargin=*, left=0pt]

\noindent
\colorchip{1f77b4}\,\,\textbf{Cluster 1} (Blue) corresponds to denser urban areas, with a sharp reliance on recent precipitation. This short lag time aligns with how impervious surfaces accelerate runoff yield~\cite{liao_fast_2023}. Furthermore, it shows higher importance of lower precipitation than others, signaling its relative vulnerability to lower precipitation and relies on relatively recent soil moisture with emphasis on the higher soil moisture values.

\noindent
\colorchip{ff7f0e}\,\,\textbf{Cluster 2} (Orange) corresponds to suburban areas in inland regions of CONUS, with moderate reliance on recent precipitation. Soil moisture kernel is similar to cluster 1, with a slightly even emphasis on all soil moisture values.

\noindent
\colorchip{2ca02c}\,\,\textbf{Cluster 3} (Green) corresponds to suburban areas along the Gulf, with a longer tail of precipitation kernel and a longer and wider lookback range than the first two clusters. Its characteristics lie in the middle of the clusters 2 and 4 drawing on both the suburban and water-facing characteristics.

\noindent
\colorchip{d62728}\,\,\textbf{Cluster 4} (Red) corresponds to water-facing regions. Represented by the longest tail in the precipitation kernel and the widest lookback range in the soil moisture kernel, this cluster also places the most even importance in soil moisture values. This may suggest water-facing regions are more susceptible to flooding even when the top layer of soil is not saturated due to the high water table.

We emphasize that the value of this result lies not in the clusters aligning with geography.
Such alignment is unsurprising, since the modulators are conditioned on foundation-model embeddings and terrain descriptors that already encode geographic structure.
It lies instead in the fact that the \textbf{clusters admit a coherent hydrological interpretation}, the architecture-level interpretability we set out to establish, which allows the physical fidelity of the foundation-model conditioning to be assessed directly from the learned parameters.
More broadly, \textbf{we view this conditioning scheme as not unique to pluvial flooding but applicable to other geospatial tasks that build on Geospatial Foundation Models}, where routing embeddings through physically meaningful parametric modules can make the model's use of them directly inspectable.
The full clustering results are placed in the Appendix.

\vspace{-6pt}
\subsection{Failure Modes}
\label{ssec:failure-modes}
% We observe two recurring main failure modes.
     \paragraph{Low-damage claims are harder to predict.} These claims are often associated with less anomalous hydrometeorological conditions, making them more difficult to distinguish from non-damage cases. 
        To characterize this quantitatively, we stratify held-out claim-days by claim amount (Table~\ref{tab:damage-severity}). 
        PR-AUC rises sharply from the smallest bin to the [\$50K, \$500K) tier ($0.04\to0.37$), with \$-Weighted PR-AUC tracking the same trend. The slight drop in the largest [\$500K, $\infty$) tier reflects its extreme base rate (0.004\%) rather than degraded ranking. 
        \textbf{DELUGE is therefore strongest in the high-cost regime most relevant to insurance and emergency management stakeholders}.
        % while underperforming in the low-cost regime.
\begin{table}[h]
\centering
\small
\setlength{\tabcolsep}{4pt}
\begin{tabular}{l r r r}
\toprule
\textbf{Damage range (\$)} & \textbf{Prev. (\%)} & \textbf{PR-AUC} & \textbf{\$-W PR-AUC} \\
\midrule
{}[0, 5{,}000)              & 0.124  & 0.043 & 0.049 \\
{}[5{,}000, 50{,}000)       & 0.087  & 0.154 & 0.186 \\
{}[50{,}000, 500{,}000)     & 0.030  & 0.368 & 0.424 \\
{}[500{,}000, $\infty$)     & 0.004  & 0.313 & 0.331 \\
\bottomrule
\end{tabular}
\caption{\textbf{DELUGE performance stratified by claim amount.} Per-bin prevalence, PR-AUC, and dollar-weighted PR-AUC on a single held-out spatial split.}
\label{tab:damage-severity}
\end{table}

\begin{figure}[]
\centering
\setlength{\tabcolsep}{1pt}
\renewcommand{\arraystretch}{0.2}
\begin{tabular}{c@{\hspace{1pt}}cccc}
\rotatebox{90}{\,\textbf{NFIP Claims}} &
\includegraphics[width=0.24\columnwidth]{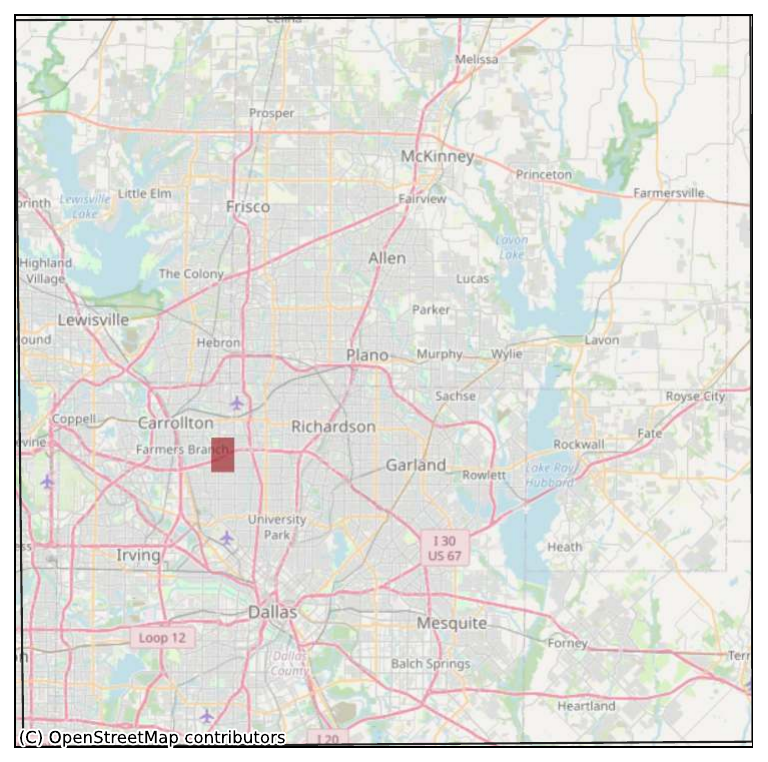} &
\includegraphics[width=0.24\columnwidth]{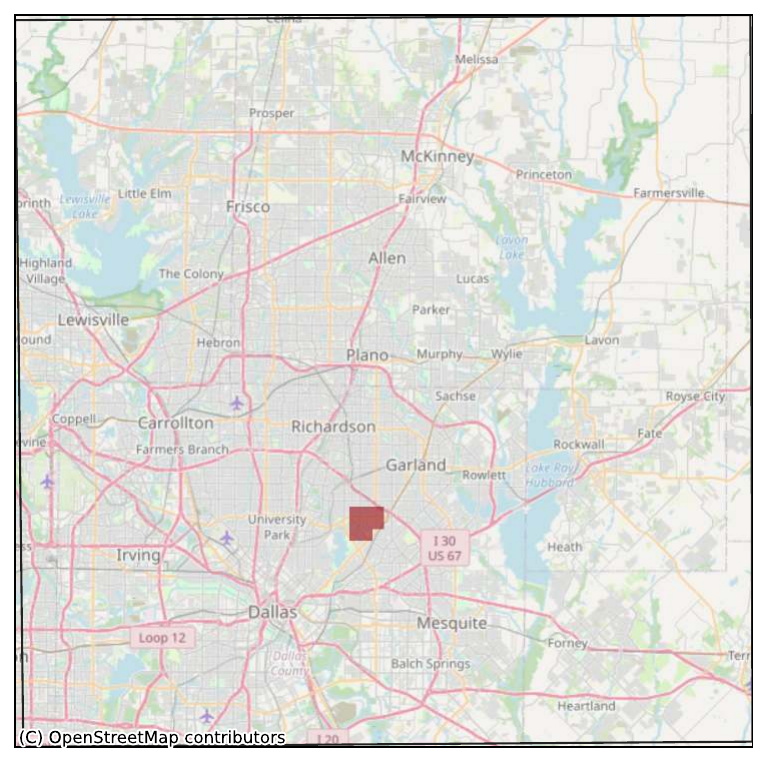} &
\includegraphics[width=0.24\columnwidth]{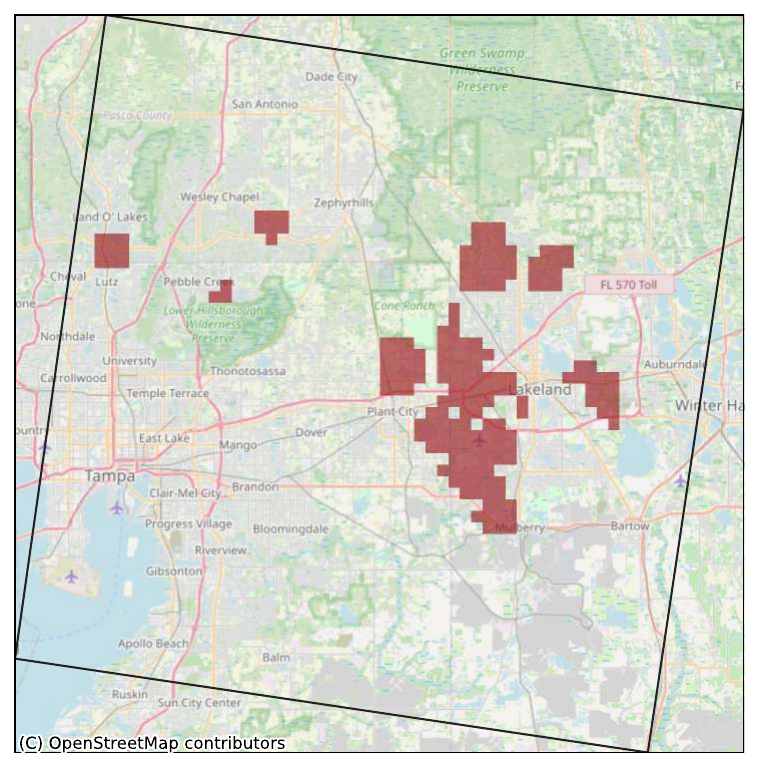} &
\includegraphics[width=0.24\columnwidth]{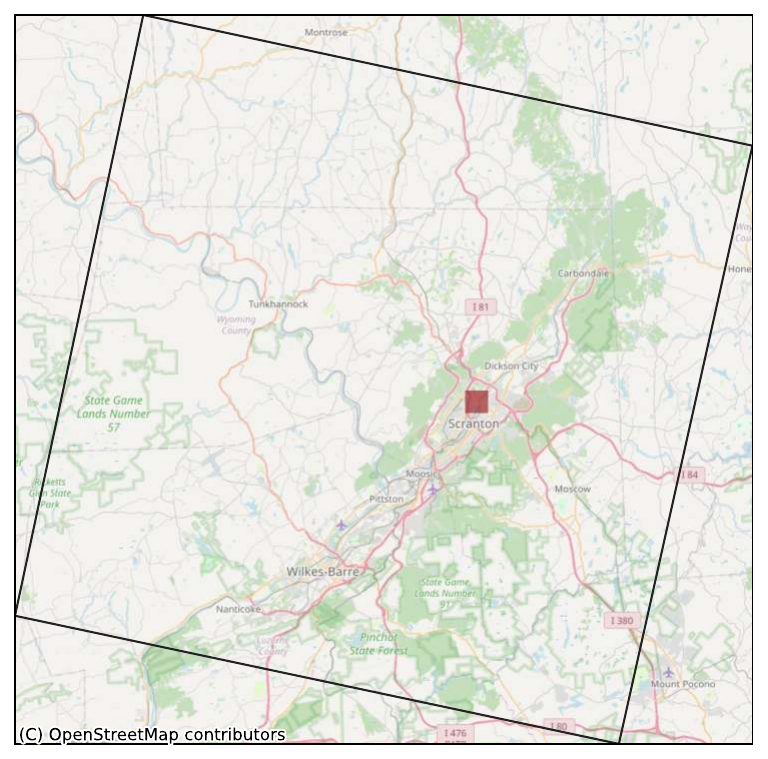} \\
\rotatebox{90}{\,\textbf{Predicted}} &
\includegraphics[width=0.24\columnwidth]{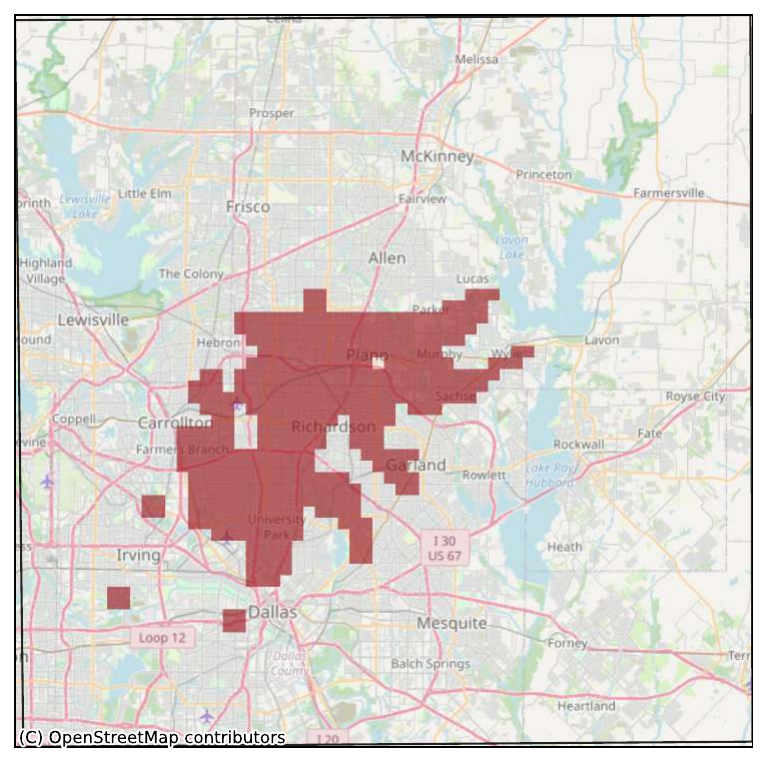} &
\includegraphics[width=0.24\columnwidth]{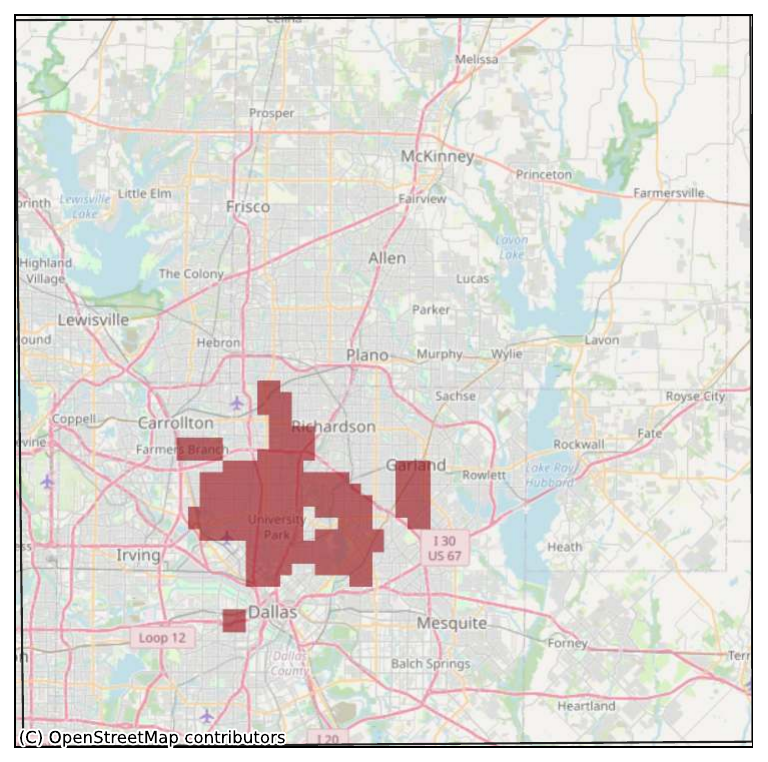} &
\includegraphics[width=0.24\columnwidth]{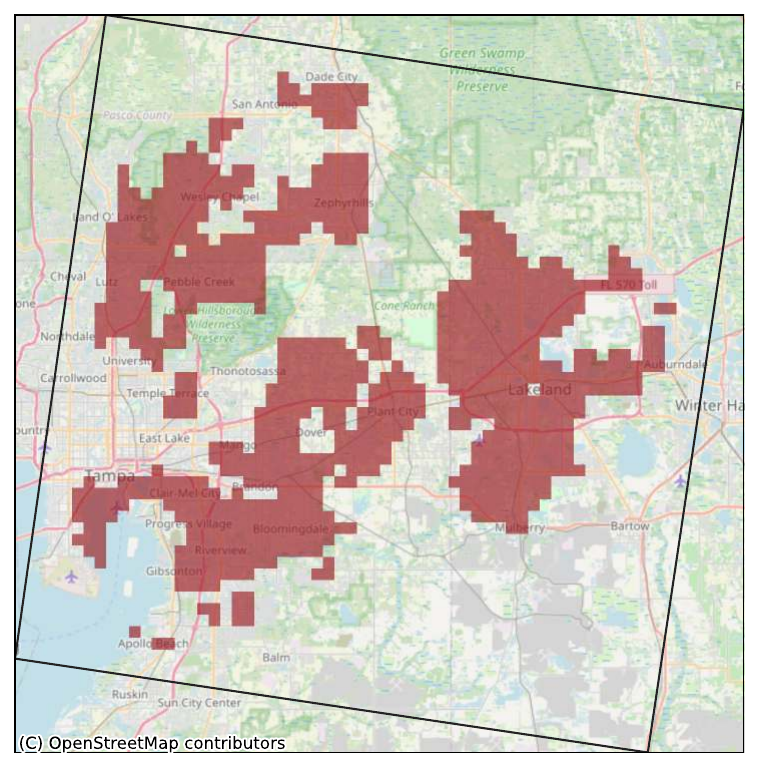} &
\includegraphics[width=0.24\columnwidth]{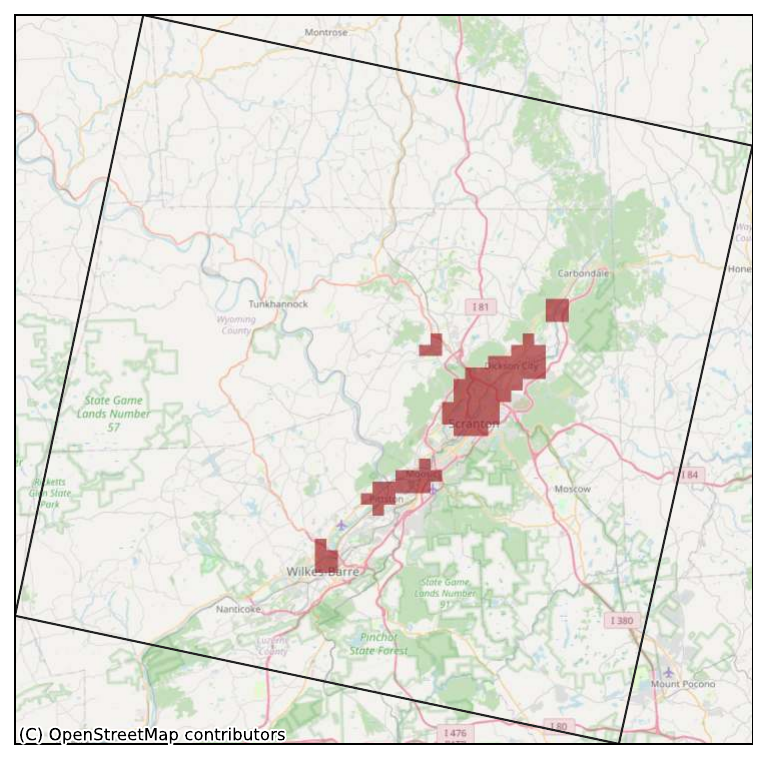} \\
                                                               % & (a) & (b) & (c) & (d)\\
\end{tabular}
\caption{DELUGE dense urban failure cases NFIP claims and DELUGE predictions  on four held-out test days where DELUGE fails to pinpoint flood claims in urban areas.}
\label{fig:failure-modes}
\end{figure}
% \vspace{-5pt}
\paragraph{Locations of impacts of large storm systems with widespread precipitation are hard to resolve.}
% \paragraph{Large storm systems with widespread precipitation are hard to localize in their impact.}
%
Pinpointing which pixels experience damage is difficult, especially in dense urban areas where flood response is governed by fine-scale local processes.
Figure~\ref{fig:failure-modes} shows representative cases where DELUGE registers that a large storm system causes damage but struggles to localize exactly which urban pixels are affected.
We attribute this to two factors.
First, because we predict only \emph{insured} damage reported through NFIP, some pixels in our negative class may have experienced unreported damage, blurring the supervisory signal.
Second, and independent of this, dense urban areas are where we most consistently observe DELUGE to struggle, as we show by stratifying performance over AEF-defined location characteristics in the Appendix.
We anticipate that higher spatial (sub-kilometer) and temporal (sub-hourly) resolution hydrometeorology will be needed to resolve fine-scale urban damage.

\vspace{-0.28in}
\section{Discussion and Conclusion}
We presented DELUGE, a multimodal CNN-based system for daily pluvial insured flood damage prediction at ${\sim}1$\,km resolution across the highest-claim regions of CONUS, trained on NFIP claims.
We structured the model around the core disaster-management components of hazard, exposure, and vulnerability, introduced an interpretable conditioning scheme for integrating foundation-model embeddings into the prediction task, and inspected the physical fidelity of that scheme.
As a prerequisite for using NFIP claims as a daily training signal, we also introduced precipitation-guided temporal and intersection-based spatial corrections that refine the timing and footprint of each claim.
DELUGE outperforms tuned gradient-boosted tree baselines on PR-AUC and concentrates high-dollar claims more effectively near the top of its prediction ranking, the regime most relevant to insurance and emergency-management stakeholders.
For insurers, these gains can improve risk segmentation. 
When calibrated to average annual loss, a model that better identifies high-dollar pluvial flood risk could support territory rating, risk selection, and reinsurance analysis, while also providing clearer risk signals for policyholders and communities.
% with the largest gains concentrated in the high-cost claim 
%
We further demonstrated the efficacy of our \emph{Value and Temporal Modulators}, which gain \emph{interpretability-by-design} while remaining competitive with standard encoders for spatial time series. 
We believe this offers a scheme transferable to other geospatial tasks built on nascent Geospatial Foundation Models.

Three limitations point to natural next steps.
First, data quality remains the dominant bottleneck, on both the label and input sides.
On the label side, NFIP claims, despite our spatial and temporal uncertainty corrections, remain a noisy and incomplete proxy for pluvial damage.
Uninsured losses and unreported events are systematically absent from the record, and claimed damage amount is itself an imperfect proxy for severity or societal importance.
The resulting sparsity in both space and time is reflected in the modest absolute performance of all models including DELUGE, which nonetheless remains strong when the low prevalence of claims is taken into account.
On the input side, our hydrometeorology inputs themselves carry nontrivial uncertainty at the hourly kilometer scale.
Improving the fidelity of both the damage signal and the hydrometeorological hazard inputs is likely the single largest lever for further predictive gains.
Second, interpretability remains open.
Our conditioning scheme is a step forward in that it exposes \emph{how} the model utilizes foundation-model embeddings, but it does not explain individual predictions.
Closing this gap, from architecture-level fidelity to case-level explanation, is a natural direction for future work.
Third, our spatial evaluation shows that DELUGE generalizes across the wide range of flood dynamics present within highest-claim regions in CONUS~\cite{brunner_spatial_2020}, but it does not yet test extrapolation to geographies absent from training.
Applying DELUGE to regions or countries outside its training distribution, where flood dynamics and the built environment may differ from anything it has seen, is a distinct problem that we leave to future work.

% We hope DELUGE serves as a startiriven pluvial flood impact modeling, and that the interpretable conditioning pattern generalizes to other geospatial prediction tasks built on foundation-model embeddings.
Together, these results suggest that DELUGE can serve as a starting 
point for CONUS-scale pluvial flood impact modeling, and that 
interpretable deep learning can turn geospatial foundation-model 
embeddings into accurate and physically meaningful prediction 
systems.

\bibliographystyle{ACM-Reference-Format}
\bibliography{template}
\clearpage
\appendix
\section{Data Sources}
\label{app:data-sources}
This appendix gives per-source descriptions of the input modalities summarized in \S\ref{ssec:arch-inputs}, organized by the risk component each contributes to.

\subsection{Hazard: Hydrometeorology}
\paragraph{NOAA Analysis of Record for Calibration (AORC)}
High quality record of precipitation with sufficient temporal and spatial resolution is critical to account for flood damages.
Past works have employed a variety of precipitation records including NASA's IMERG~\cite{huffman_integrated_2020} or ERA5~\cite{hersbach_era5_2020} reanalysis data; however, these datasets are insufficient for the daily, $\sim$1km scale of our analysis.
In this work, we use the NOAA Analysis of Record for Calibration (AORC) dataset~\cite{fall_analysis_2023}, which provides a high quality, high resolution (hourly, $\sim$800m) precipitation record for the Conterminous United States.
We use the \textbf{\textit{Hourly Accumulated Precipitation}} record from this archive.

\paragraph{NOAA National Water Model (NWM) Retrospective}
Precipitation signal alone is often insufficient to model pluvial flood occurrence and intensity, as past works have shown that pluvial flooding is fundamentally a hydrological problem~\cite{james_precipitation_2024}.
Thus to supplement the NOAA AORC precipitation records, we turn to NOAA's National Water Model (NWM) Retrospective dataset v3.0~\cite{cosgrove_noaas_2024} available from Feb. 1979 to Jan. 2023.
Amongst its various outputs, we use the Land Surface Model component, which provides 3-hourly records of physical land surface and hydrologic states at $\sim$1\,km resolution.
We use two variables from the NWM Retrospective dataset: \textbf{\textit{Volumetric Soil Moisture}} and \textbf{\textit{Snow Water Equivalent}}.

\subsection{Hazard: Terrain}

\paragraph{Topographic Wetness Index (TWI)}
\textbf{\textit{Topographic Wetness Index}} quantifies the steady-state propensity of a location to accumulate surface water from its upslope contributing area, defined as $\mathrm{TWI} = \ln(a/\tan\beta)$, where $a$ is the upslope contributing area per unit contour length and $\beta$ is the local slope.
We use the CONUS-wide 30m product from~\cite{hoylman_30m_2021}.

\paragraph{Height Above Nearest Drainage (HAND)}
\textbf{\textit{Height Above Nearest Drainage}} measures the vertical elevation of each cell above its hydrologically connected drainage network.
Low-HAND cells lie nearest the drainage corridors that inundate first when local drainage capacity is exceeded, making HAND a direct proxy for pluvial flood susceptibility.
We use the Copernicus DEM HAND dataset at 30\,m resolution~\cite{aws_copernicus_dem}.

\paragraph{Global Curve Number (GCN)}
The SCS Curve Number parameterizes surface runoff potential as a joint function of soil hydrologic group and land cover, with values ranging from $\sim$30 (high infiltration on sandy or forested terrain) to 100 (fully impervious surfaces).
Higher CN indicates a larger fraction of incident precipitation converted to surface runoff, the proximate hydrologic driver of pluvial flooding.
CN values from~\cite{jaafar_gcn250_2019} are derived from three different antecedent runoff conditions (ARC): dry (ARC I), average (ARC II), and wet (ARC III) soil moisture states, which we include as separate features.
We use the three \textbf{\textit{Global Curve Numbers}} as terrain hazard descriptors.

\paragraph{NLCD Land Cover}
Land cover plays an important role in pluvial flood risk with impervious surfaces generating more runoff and thus more pluvial flood risk.
To account for impervious surfaces, we include the National Land Cover Database (NLCD) 2016 product~\cite{usgs_annual_nlcd_2024}, which provides 30m land cover classifications across the Conterminous United States.
We use the \textbf{\textit{Fractional Impervious Surface}} classification as a surface descriptor.

\paragraph{POLARIS Soil Properties}
Soil plays a critical role in modeling pluvial flood events~\cite{rong_impact_2024}, as it controls the partitioning of precipitation into infiltration and runoff.
To account for the varied soil properties across the Conterminous United States, we include the POLARIS dataset~\cite{chaney_polaris_2019}, which provides high-resolution (30m) estimates of soil properties such as texture, hydraulic conductivity, and water retention characteristics.
From the dataset, we identify three variables of interest: $K_{sat}$, the \textbf{saturated hydraulic conductivity} (\emph{$\log_{10}(cm/hr)$}); \emph{clay}, the \textbf{clay percentage} (\%); and $\theta_{s}$, the \textbf{saturated soil water content} ($m^3/m^3$).
All values are derived from the 0--5\,cm soil layer.

\subsection{Exposure and Vulnerability}
\paragraph{FIMA NFIP Policies}
The FIMA NFIP policies dataset provides spatially aggregated counts and total coverage of active flood insurance policies across the Conterminous United States.
We use policy density and total insured value as exposure proxies; regions with denser NFIP coverage contain more structures with known flood vulnerability and a larger insured asset base at risk.
These are derived from the OpenFEMA FIMA Redacted Policies Dataset v2~\cite{fema_nfip_policies_v2} and undergo the same spatial uncertainty correction as the claims data (\S\ref{ssec:spatial_correction}), yielding a polygon-level count and total coverage for each cell in our analysis grid.
We derive \textbf{\textit{\# Active NFIP Policy}} from this archive.

\paragraph{National Structure Inventory (NSI)}
The U.S. Army Corps of Engineers National Structure Inventory~\cite{usace_nsi_2022} provides a point-level inventory of structures across CONUS, with per-structure attributes including replacement value, occupancy type, square footage, first-floor elevation, and foundation type.
From this archive we derive and rasterize using the same methodology as in \S\ref{ssec:spatial_correction} the following variables: \textbf{Total Count of Structures}, \textbf{\% Structs with Basements}, \textbf{\% Struct. with Slab Foundation}, \textbf{\% Mobile Homes}, \textbf{\% Structs in Special Flood Hazard Area (SFHA)}, and \textbf{Average, Min, Standard Deviation First Floor Elevation}.

\subsection{Google AlphaEarth Foundation (AEF)}
The AlphaEarth Foundation model provides a 64-dimensional learned embedding of Earth's surface at 10\,m resolution, trained on multi-source Earth observation imagery and other environmental datasets~\cite{brown_alphaearth_2025}.
AEF embeddings implicitly encode land cover, impervious surface fraction, urbanization patterns, vegetation, and built-environment morphology, all factors that modulate runoff generation and drainage capacity, providing a complementary, dense representation alongside the hand-engineered terrain and exposure features above.

\subsection{Normalization}
To stabilize training, we apply transformations to our raw feature values.
For those that exhibit high right skew, \textbf{HAND}, \textbf{TWI}, \textbf{\# Active NFIP Policies}, \textbf{\# NSI Structures}, we apply a $\ln(x+1)$ transformation followed by a z-score normalization.
For \textbf{NSI Foundation height features}, we apply a z-score normalization without the log transform, as these features are not as heavily skewed.
For the rest we apply a min-max normalization to scale values to the [0,1] range.
\section{Date Shift Distribution}
We describe our temporal correction procedure in \S\ref{ssec:temporal_correction}. 
Here, we report the distribution of date shifts $\Delta = d' - d$ applied to the original NFIP claim dates $d$ to obtain the corrected dates $d'$ used in our study region. As shown in Table~\ref{tab:date_shifts}, the vast majority of claims ($81.1\%$) had no date shift, while $7\%$ has shifts of $\pm 1$ day and $6\%$ were discarded.
\begin{table}[h]
\small
\centering
\setlength{\tabcolsep}{5pt}
\begin{tabular}{r *{8}{c}}
\toprule
$\Delta$ (days) & $-3$ & $-2$ & $-1$ & $\phantom{-}0$ & $+1$ & $+2$ & $+3$ & -- \\
\midrule
\% & $2.0$ & $1.9$ & $5.0$ & $\mathbf{81.1}$ & $2.0$ & $1.4$ & $0.6$ & $6.0$ \\
\bottomrule
\end{tabular}
\caption{Distribution of date shifts $\Delta = d' - d$ from the temporal correction. (See \S\ref{ssec:study_region})}
\label{tab:date_shifts}
\end{table}

\section{All Held-Out Cell Modulator Dynamics}
\label{app:all-cell-dynamics}
\begin{figure*}[!ht]
\centering
\setlength{\tabcolsep}{1pt}
\renewcommand{\arraystretch}{0.4}
\newcommand{\cellbasemap}[1]{\includegraphics[width=0.15\textwidth]{figs/cell_dynamics_val_cell_#1_basemap.pdf}}
\begin{tabular}{cccccc}
\cellbasemap{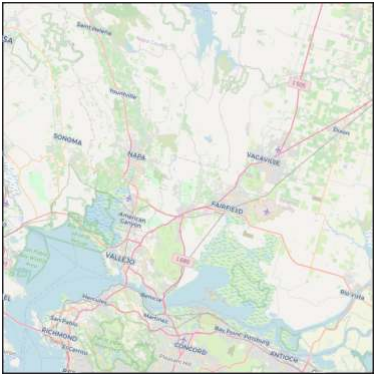}   & \cellbasemap{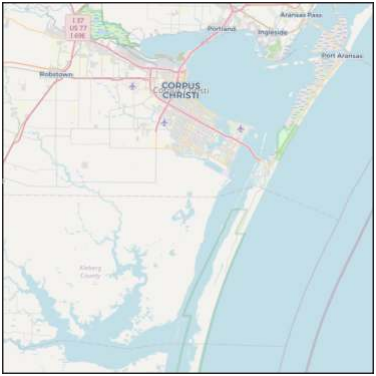} & \cellbasemap{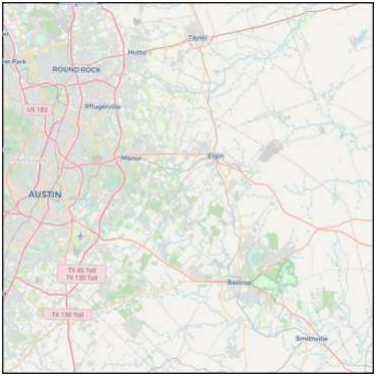} & \cellbasemap{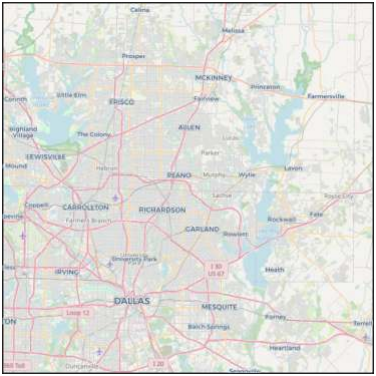} & \cellbasemap{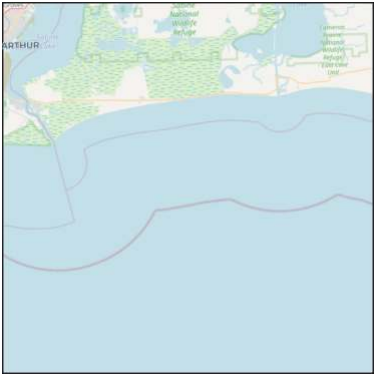} & \cellbasemap{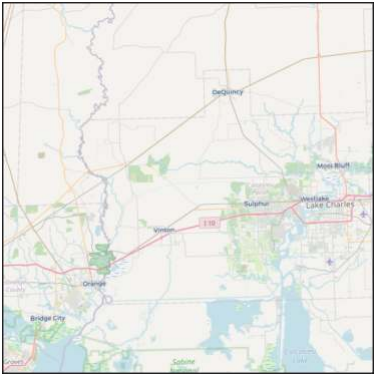} \\
\cellbasemap{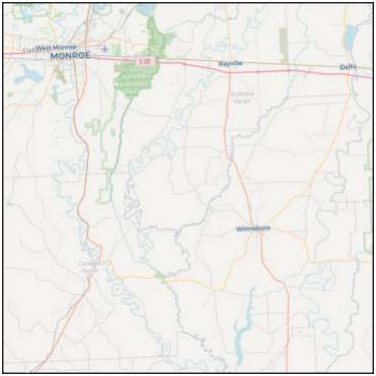} & \cellbasemap{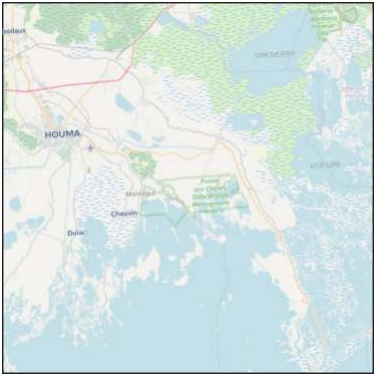} & \cellbasemap{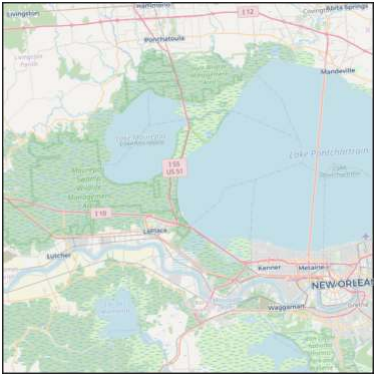} & \cellbasemap{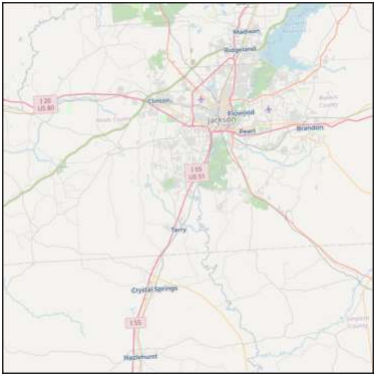} & \cellbasemap{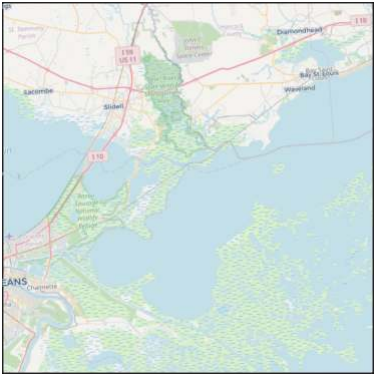} & \cellbasemap{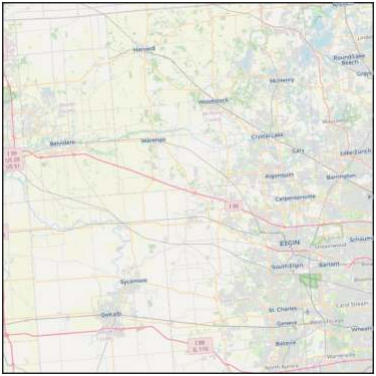} \\
\cellbasemap{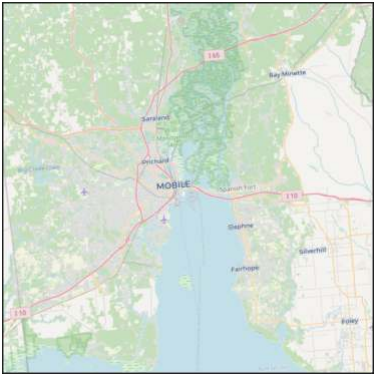} & \cellbasemap{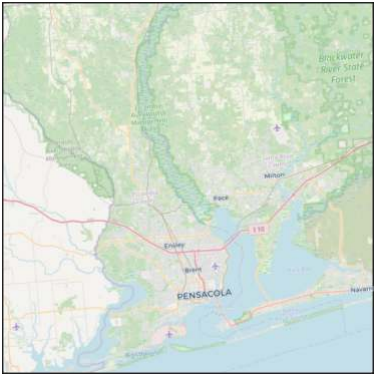} & \cellbasemap{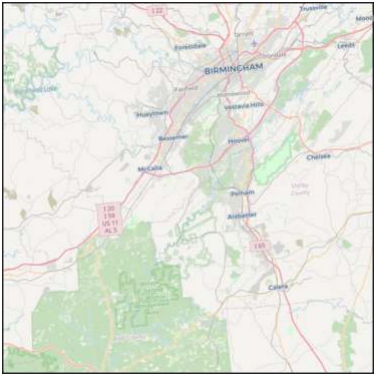} & \cellbasemap{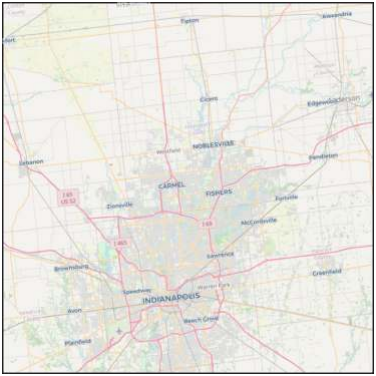} & \cellbasemap{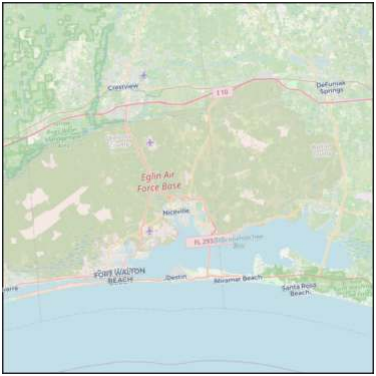} & \cellbasemap{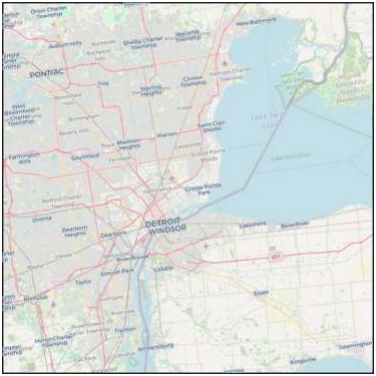} \\
\cellbasemap{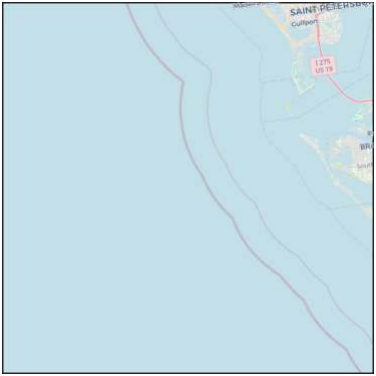} & \cellbasemap{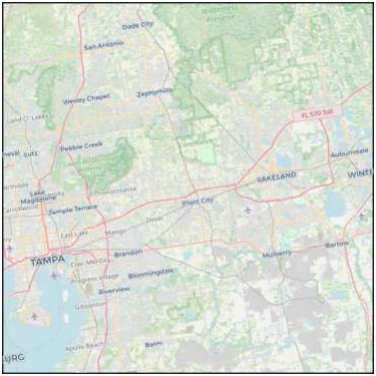} & \cellbasemap{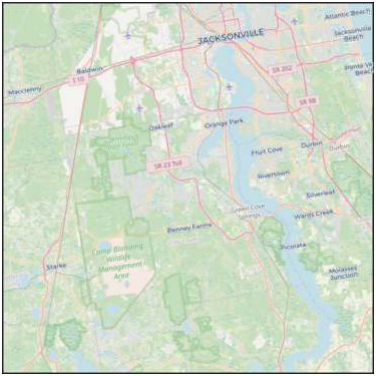} & \cellbasemap{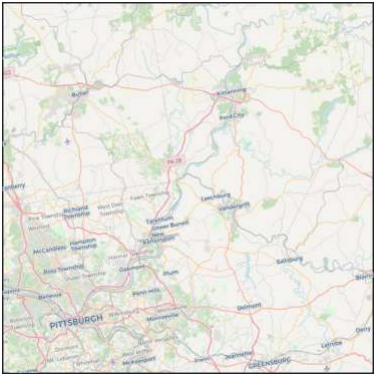} & \cellbasemap{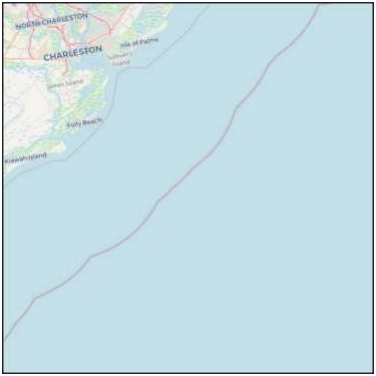} & \cellbasemap{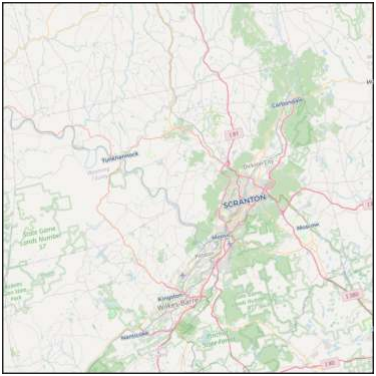} \\
\cellbasemap{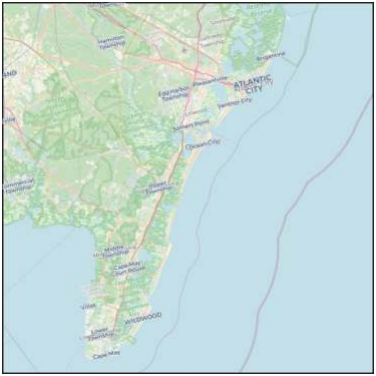} & \cellbasemap{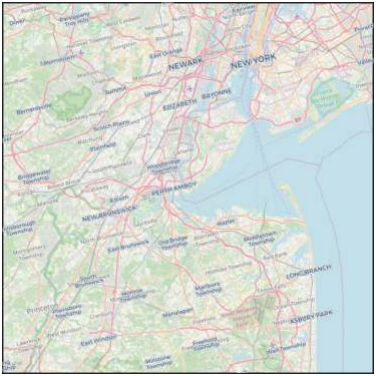} & \cellbasemap{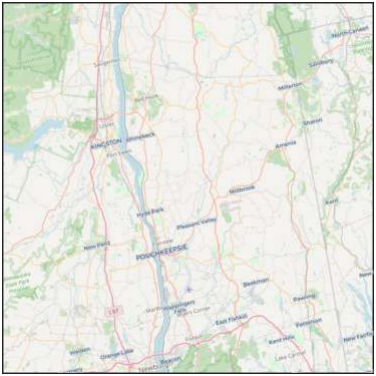} & \cellbasemap{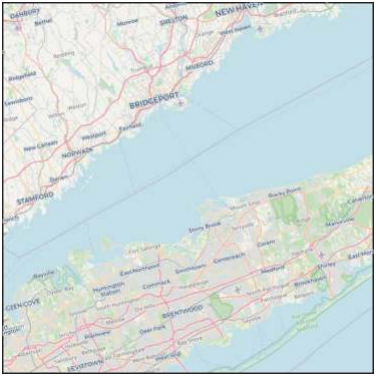} & \cellbasemap{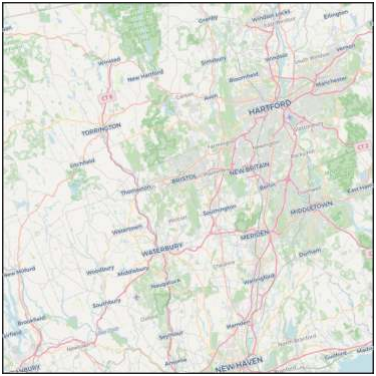} & \cellbasemap{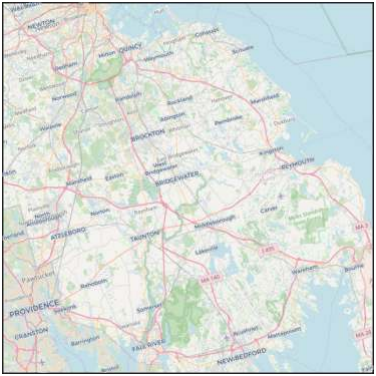} \\
\end{tabular}
\caption{Basemaps of all 30 held-out cells, paired with the modulator kernels in Figure~\ref{fig:all-cell-kernels}}
\label{fig:all-cell-basemaps}
\end{figure*}

\begin{figure*}[!ht]
\centering
\includegraphics[width=\textwidth]{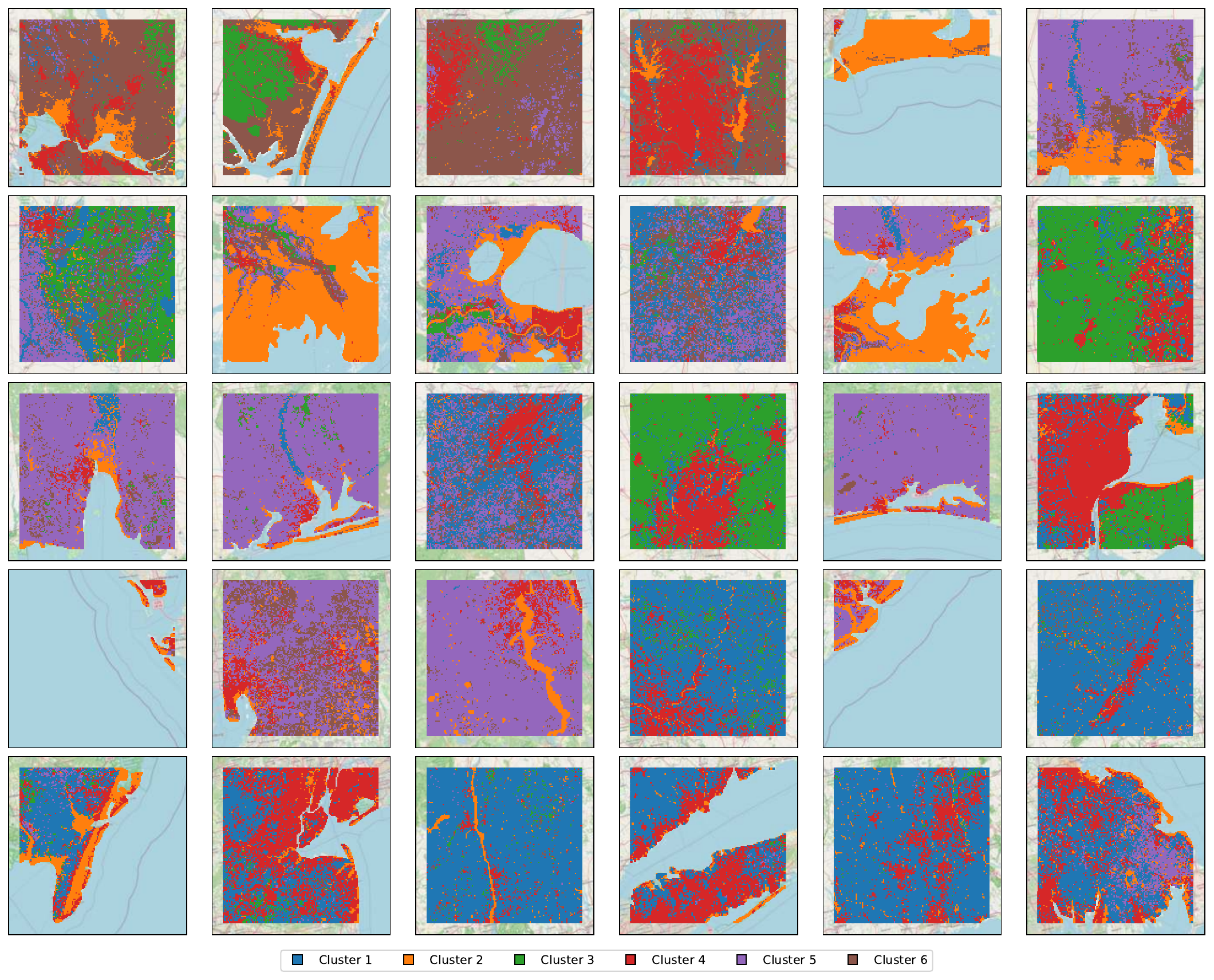}
\caption{Google AlphaEarth Clustering results for all 30 held-out cells. See \S\ref{ssec:loc_char}.}
\label{fig:all-aef-cluster}
\end{figure*}

\begin{figure*}[!ht]
\centering
\setlength{\tabcolsep}{1pt}
\renewcommand{\arraystretch}{0.4}
\newcommand{\cellkernel}[1]{\includegraphics[width=0.15\textwidth]{figs/cell_dynamics_val_cell_#1.pdf}}
\begin{tabular}{cccccc}
\cellkernel{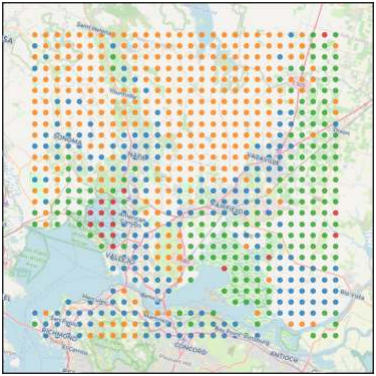}   & \cellkernel{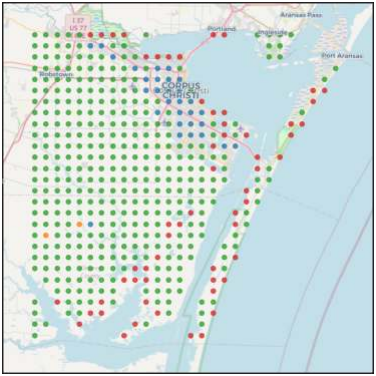} & \cellkernel{1136} & \cellkernel{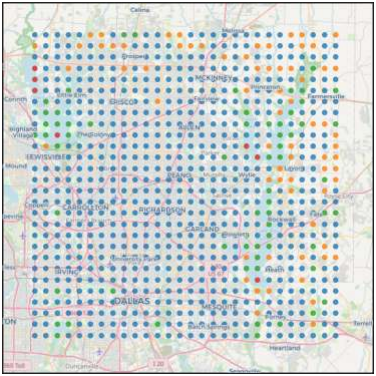} & \cellkernel{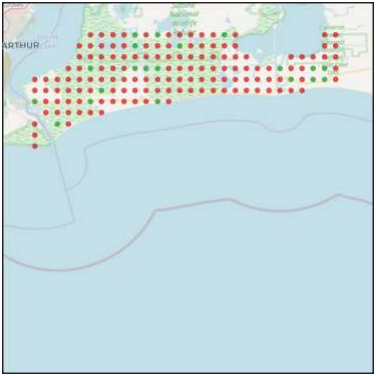} & \cellkernel{1331} \\
\cellkernel{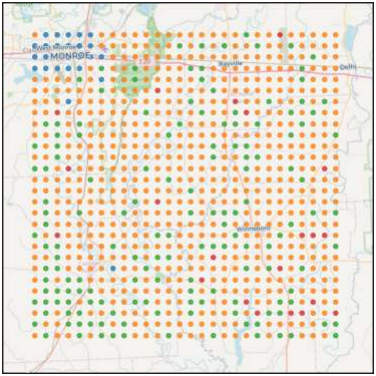} & \cellkernel{1486} & \cellkernel{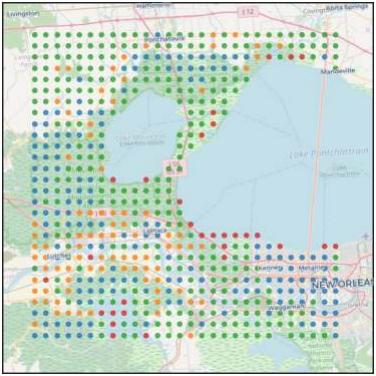} & \cellkernel{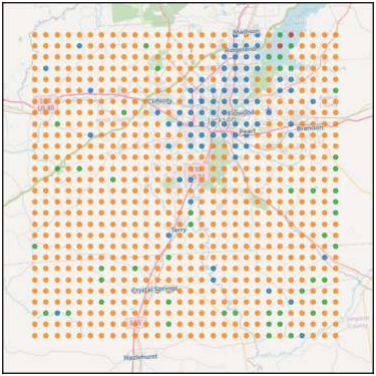} & \cellkernel{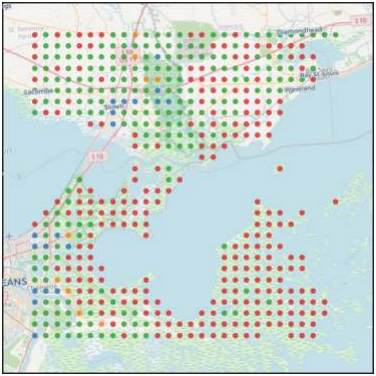} & \cellkernel{1544} \\
\cellkernel{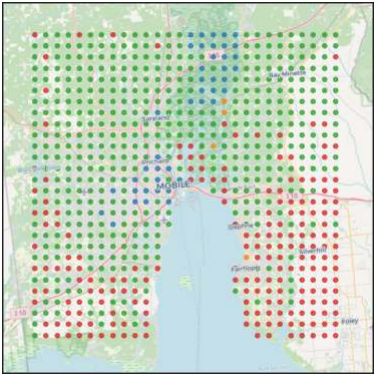} & \cellkernel{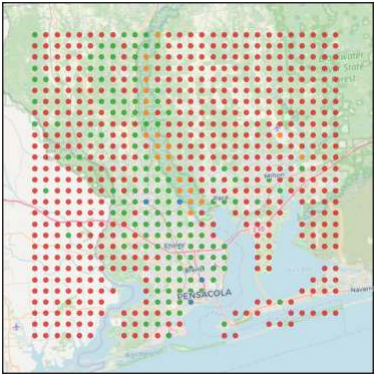} & \cellkernel{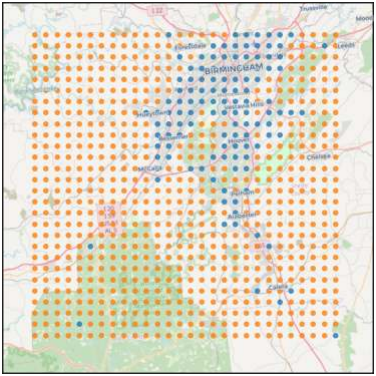} & \cellkernel{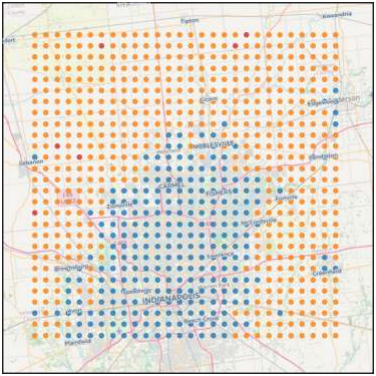} & \cellkernel{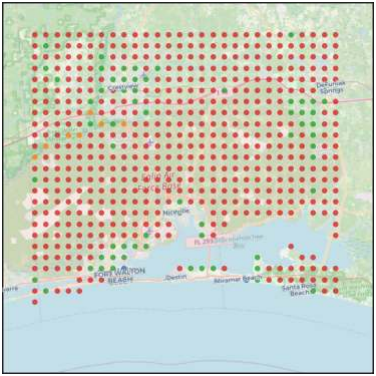} & \cellkernel{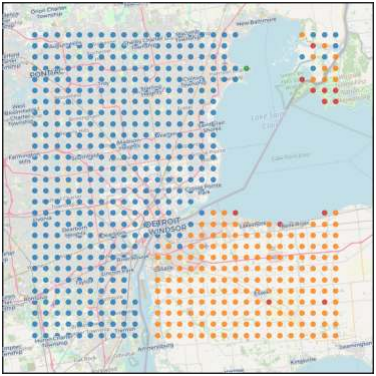} \\
\cellkernel{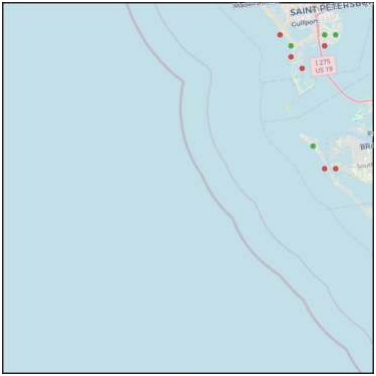} & \cellkernel{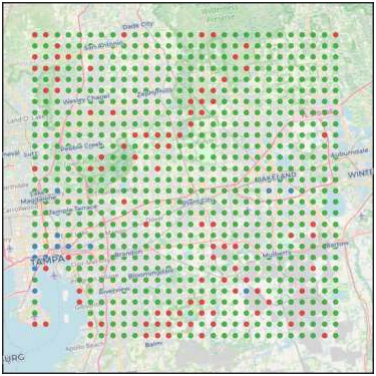} & \cellkernel{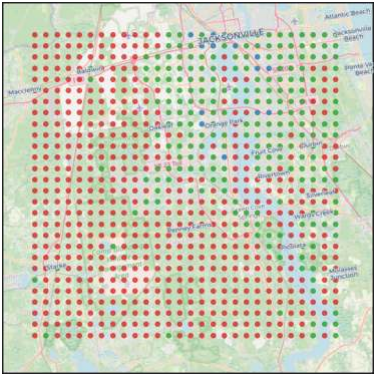} & \cellkernel{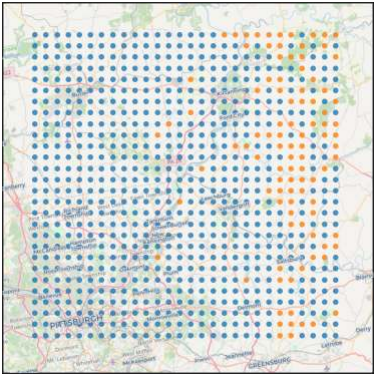} & \cellkernel{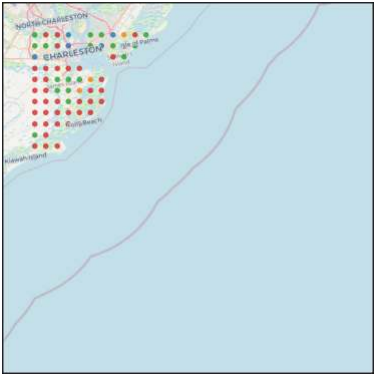} & \cellkernel{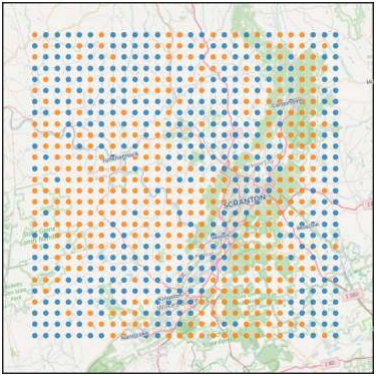} \\
\cellkernel{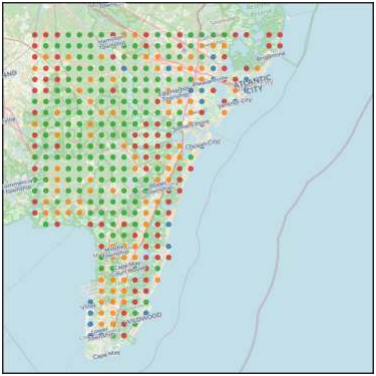} & \cellkernel{2168} & \cellkernel{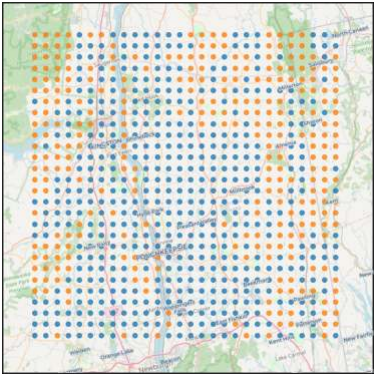} & \cellkernel{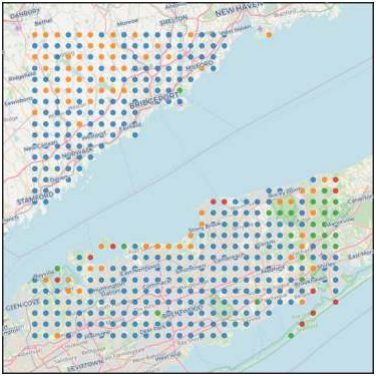} & \cellkernel{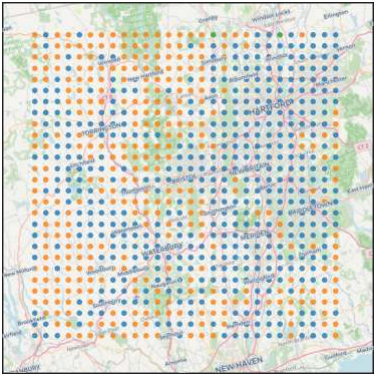} & \cellkernel{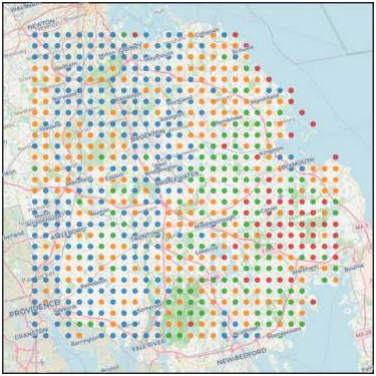} \\
\end{tabular}
\caption{Learned modulator clustering for all 30 held-out cells. See \S\ref{ssec:modulator_behaviors}. }
\label{fig:all-cell-kernels}
\end{figure*}

For completeness, Figures~\ref{fig:all-cell-basemaps},~\ref{fig:all-aef-cluster} and~\ref{fig:all-cell-kernels} show, respectively, the basemap, the AEF clustering, and the learned modulator clustering for every held-out cell. Panel positions match across the three figures.


\section{Temporal Generalization}
Under a strict temporal split (train 2017--2020, test 2021--2022, single split), DELUGE retains its lead on both PR-AUC and \$-Weighted PR-AUC (Table~\ref{tab:temporal}).
PR-AUC degrades only modestly from the spatial split, with DELUGE the least affected at a \tsim$3\%$ drop and the tree baselines down \tsim$7$ to $10\%$, preserving DELUGE's lead and the baseline ordering.
\$-Weighted PR-AUC drops sharply for every model (roughly \tsim$38$ to $41\%$ from the spatial split), but DELUGE retains its lead and the ordering across baselines is preserved.
\begin{table}[h]
\small
\centering
\setlength{\tabcolsep}{4pt}
\begin{tabular}{l r r r r}
\toprule
\textbf{Model} & \textbf{PR-AUC} & \textbf{$\Delta$} & \textbf{\$-W PR-AUC} & \textbf{$\Delta$} \\
\midrule
\textbf{DELUGE (Ours)} & $\mathbf{0.237}$ & --      & $\mathbf{0.370}$ & --      \\
XGBoost                & $0.197$          & $-$17\% & $0.303$          & $-$18\% \\
LightGBM               & $0.212$          & $-$11\% & $0.329$          & $-$11\% \\
\bottomrule
\end{tabular}
\caption{\textbf{Temporal split (train 2017--2020, test 2021--2022, single split).} DELUGE retains its lead on both metrics.}
\label{tab:temporal}
\end{table}
The divergence between metrics is structural. 
PR-AUC averages over all positives whereas \$-Weighted PR-AUC concentrates on the few costliest claims, so models are judged primarily on how well they rank the tail.
% %
% And the tails of the two windows differ in obviousness. 
%
Training is dominated by Hurricane Harvey in 2017 (\$5.8B, more than the other three training years combined at \$2.0B), a clear signal that any reasonable model can learn to flag. 
Test years (\$1.4B in 2021, \$1.9B in 2022) lack a comparably obvious extreme, which hampers the \$-Weighted PR-AUC of all models.

\section{Performance Comparison by Location Characteristics}\label{ssec:loc_char}
To understand whether DELUGE's headline performance is uniform across our study cells or concentrated in particular locations, we stratify the held-out cells by their underlying landscape and report PR-AUC within each stratum.
We characterize each held-out pixel by its AlphaEarth Foundation embedding~\cite{brown_alphaearth_2025} and cluster the resulting per-pixel vectors with K-means and a silhouette sweep selects $K=6$.
We then report DELUGE's PR-AUC within each AEF cluster (Table~\ref{tab:loc-char}), along with the \emph{lift over prevalence} (PR-AUC $/$ prevalence), the multiplicative gain over a random predictor.
Inspecting the clusters' geographic membership, we label them as seen in Table~\ref{tab:loc-char}.
% : \textbf{C1} Northeast developed areas, \textbf{C2} coastal wetland, \textbf{C3} inland developed areas, \textbf{C4} dense developed areas, \textbf{C5} Southeast mixed rural-developed, and \textbf{C6} inland less-developed areas.
%
DELUGE's raw PR-AUC varies substantially across these strata.
Lift over prevalence largely controls for this, and across most clusters the gain over chance is comparable.

Of note, however, is the poor relative lift of the \textbf{dense developed} cluster (C4, $42\times$).
Despite the highest raw PR-AUC of any built-environment stratum, DELUGE's marginal contribution over a base-rate predictor is the smallest here.
We attribute this to \textbf{local, sub-kilometer flood dynamics in dense urban areas, that operate below the ${\sim}1$\,km resolution at which DELUGE operates.}
This matches the localization failure mode to be discussed in \S\ref{ssec:failure-modes} and points to spatial resolution as a natural lever for future work in this regime.
We place the full cluster membership and geographic distribution in Appendix.

\begin{table}[t]
\small
\centering
\setlength{\tabcolsep}{4pt}
\begin{tabular}{l r r r}
\toprule
\textbf{Cluster}  & \textbf{Prev. (\%)} & \textbf{PR-AUC} & \textbf{Lift} \\
\midrule
C1: NE developed             & $0.139$ & $0.184$         & $132{\times}$ \\
C2: Coastal wetland          & $0.166$ & $0.138$         & $83{\times}$ \\
C3: Inland developed         & $0.065$ & $0.080$         & $124{\times}$ \\
C4: Dense developed          & $0.610$ & $0.255$         & $42{\times}$ \\
C5: SE mixed rural-developed & $0.249$ & $0.322$ & $129{\times}$ \\
C6: Inland less-developed    & $0.222$ & $0.238$         & $107{\times}$ \\
\bottomrule
\end{tabular}
\caption{DELUGE performance stratified by AEF cluster. Held-out cells grouped by K-means ($K=6$) on the center-pixel AlphaEarth embedding. Prev.\ is per-pixel-day prevalence and lift is PR-AUC $/$ prevalence, the multiplicative gain over a random predictor.}
\label{tab:loc-char}
\end{table}

\end{document}